%% file: main.tex
\documentclass{article} % For LaTeX2e
\usepackage[algo2e,ruled,vlined,linesnumbered]{algorithm2e}
\usepackage[sglblindworkshop,final,nonatbib]{ai4nextg_neurips_2025}
\usepackage{times}
\usepackage[authoryear,round]{natbib} % FEMU
\usepackage{hyperref}
\usepackage{url}
\usepackage{subcaption}
\usepackage{wrapfig}
\usepackage{graphicx}
\usepackage{caption}
\usepackage{float}
\usepackage{booktabs}
\usepackage{makecell}
\usepackage{amssymb}
\usepackage{algpseudocode}
\usepackage{amsmath}
\usepackage{multicol}
\usepackage{xfrac}
\usepackage{xcolor}
\usepackage{cleveref}

\newcommand{\hw}[1]{\textcolor{black}{\textbf{}#1}}%brown
\newcommand{\fm}[1]{\textcolor{black}{\textbf{}#1}}%orange
\newcommand{\cameradone}[1]{\textcolor{black}{\textbf{}#1}}%green

\renewcommand{\SetKwInOut}[2]{%
  \sbox\algocf@inoutbox{\KwSty{#2}\algocf@typo:}%
  \expandafter\ifx\csname InOutSizeDefined\endcsname\relax% if first time used
    \newcommand\InOutSizeDefined{}\setlength{\inoutsize}{\wd\algocf@inoutbox}%
    \sbox\algocf@inoutbox{\parbox[t]{\inoutsize}{\KwSty{#2}\algocf@typo:\hfill}~}\setlength{\inoutindent}{\wd\algocf@inoutbox}%
  \else% else keep the larger dimension
    \ifdim\wd\algocf@inoutbox>\inoutsize%
    \setlength{\inoutsize}{\wd\algocf@inoutbox}%
    \sbox\algocf@inoutbox{\parbox[t]{\inoutsize}{\KwSty{#2}\algocf@typo:\hfill}~}\setlength{\inoutindent}{\wd\algocf@inoutbox}%
    \fi%
  \fi% the dimension of the box is now defined.
  \algocf@newcommand{#1}[1]{%
    \ifthenelse{\boolean{algocf@inoutnumbered}}{\relax}{\everypar={\relax}}%
    {\let\\\algocf@newinout\hangindent=\inoutindent\hangafter=1\parbox[t]{\inoutsize}{\KwSty{#2}\algocf@typo:\hfill}~##1\par}%
    \algocf@linesnumbered% reset the numbering of the lines
  }}%
\makeatother

\SetKwInput{KwInput}{Input}
\SetKwInput{KwOutput}{Output}

\title{MAR-FL: A Communication Efficient Peer-to-Peer Federated Learning System}

\author{
    Felix Mulitze \\
    Technical University of Munich \\
    \texttt{f.mulitze@tum.de} \And
    Herbert Woisetschl\"ager \\
    Technical University of Munich \\
    \texttt{h.woisetschlaeger@tum.de} \And
    Hans-Arno Jacobsen \\
    University of Toronto \\
    \texttt{jacobsen@eecg.toronto.edu}
}

\begin{document}

\maketitle

% TLDR: MAR-FL is a new peer-to-peer federated learning system that uses iterative group-based aggregation to reduce communication costs from O(n²) to O(n log n), making it more scalable and robust for wireless networks with unreliable participants.

\begin{abstract}
The convergence of next-generation wireless systems and distributed Machine Learning (ML) demands Federated Learning (FL) methods that remain efficient and robust with wireless connected peers and under network churn. 
Peer-to-peer (P2P) FL removes the bottleneck of a central coordinator, but existing approaches suffer from excessive communication complexity, limiting their scalability in practice. 
We introduce \textbf{MAR-FL}, a novel P2P FL system that leverages iterative group-based aggregation to substantially reduce communication overhead while retaining resilience to churn. 
MAR-FL achieves communication costs that scale as $\mathcal{O}(N\log{N})$, contrasting with the $\mathcal{O}(N^2)$ complexity of previously existing baselines, and thereby maintains effectiveness especially as the number of peers in an aggregation round grows. 
The system is robust towards unreliable FL clients and \hw{can} integrate private computing. 

% Peer-to-peer (P2P) Federated Learning (FL) eliminates the central coordinator, thereby removing communication bottlenecks and single points of failure inherent in client–server FL. 
% Yet existing P2P FL approaches remain impractical in bandwidth-constrained or large-scale environments due to excessive communication complexity and limited robustness to churn. 
% We introduce MAR-FL, a novel P2P FL system that leverages iterative group-based aggregation~\citep{ryabinin21} to significantly reduce communication costs while ensuring resilience to network disturbances. 
% Across benchmarks on MNIST and 20 Newsgroups, MAR-FL attains communication overhead that is orders of magnitude lower than established P2P FL baselines and exhibits substantially better scalability. 
% Its favorable communication complexity, scaling as $\mathcal{O}(n\log{n})$ compared to the $\mathcal{O}(n^2)$ of ring or All-Reduce baselines, enables effective training even as the number of participants grows. 
% Further experiments demonstrate that MAR-FL maintains efficiency under partial participation and integrates seamlessly with privacy-preserving mechanisms. 
% These results establish MAR-FL as a scalable and communication-efficient foundation for practical P2P FL deployments. 

% The abstract paragraph should be indented 1/2~inch (3~picas) on both left and
% right-hand margins. Use 10~point type, with a vertical spacing of 11~points.
% The word \textbf{Abstract} must be centered, bold, and in point size 12. Two
% line spaces precede the abstract. The abstract must be limited to one
% paragraph.
\end{abstract}

\vspace{-12pt}
\section{Introduction}
\vspace{-7pt}
\label{sec:intro}

\input{chapters/01_introduction}

\vspace{-7pt}
\section{MAR-FL: Communication-efficient P2P Federated Learning}
\vspace{-7pt}
\label{sec:marfl}

\input{chapters/02_marfl}

\vspace{-7pt}
\section{Experiments}
\vspace{-7pt}
\label{sec:results}

\input{chapters/03_experiments}

\vspace{-7pt}
\section{Conclusions}
\vspace{-7pt}
\label{sec:conclusions}
We introduce MAR-FL, a P2P FL system that leverages iterative group-based 
aggregation to substantially reduce communication costs compared to existing 
P2P FL techniques. 
On 125 peers, MAR-FL requires about $10\times$ less total communication than 
RDFL or AR-FL while achieving identical model utility. 
MAR-FL scales with $\mathcal{O}(N\log{N})$, enabling efficient training as the number of peers grows. 
Moreover, our system remains robust under unreliable peers, supports KD to further reduce communication, and integrates DP. 
Our findings position MAR-FL as a practical foundation for scalable, communication-efficient P2P FL in next-generation wireless settings.

\textbf{Limitations.}
\hw{While MAR-FL improves the communication efficiency of P2P FL, there is still a performance gap towards client-server FL. 
Such performance penalties cause higher operating costs, which typically hinders practical adoption.
We offer a starting point for using DP with MAR-FL but analyzing the impact of group-based aggregation in combination with momentum on DP dynamics remains open.
}

% \fm{MAR-FL moves in the right direction but does not fully close the performance gap to client–server FedAvg: reaching the same model utility still requires more communication in our fully decentralized setting. 
% We see part of this gap as intrinsic to decentralization and part as algorithmic. 
% Closing it -- especially under partial participation and network churn -- remains future work. 
% Note that, unlike FedAvg, MAR-FL avoids a single server and distributes the load over multiple rounds, which can be advantageous in unreliable networks.}

\textbf{Future work.}
\fm{Future work includes a thorough analysis of partial participation and network churn -- bringing our system even closer to real-world applicability. 
Exploring approximate aggregation and adaptive group-based information propagation could further improve communication efficiency and narrow the gap to client–server FedAvg. 
Experimental evaluations of integrating DP into MAR-FL should exploit our system’s scalability to compress peer-sampling rates; maintaining model utility while reducing the privacy loss. 
Finally, we emphasize the importance of P2P FL: by omitting a centralized server, MAR-FL avoids communication and memory bottlenecks inherent in client–server FL and moves FL closer to its promising applications.} 

% Future work includes several directions to further optimize MAR-FL.

% \textbf{Future Work.}
% Future work may address the following topics: a thorough analysis of partial participation; exploration of approximate aggregation to optimize communication efficiency; and experimental evaluations of DP integration with compressed peer-sampling rates.
% % "exploration of approximate aggregation to optimize communication efficiency"
% % "DP peer-s. compressing through scalability"
% We emphasize the importance of P2P FL: omitting a centralized coordination server removes the communication and memory bottlenecks of client–server FL and brings FL closer to its promising applications. 

% Our results demonstrate that MAR-FL leverages iterative group-based aggregation to substantially lower communication over P2P FL baselines and scales with $\mathcal{O}(n\log{n})$, enabling effective training as the number of peers grows. 
% Across \texttt{MNIST} and \texttt{20NG}, MAR-FL matches baseline accuracy while remaining robust to unreliable peers. 
% Moreover, KD can be used to further reduce total communication costs and we enable the integration of DP. 
% Taken together, our findings position MAR-FL as a practical foundation for scalable, communication-efficient P2P FL in next-generation wireless settings.
% For future work we suggest the following topics: a thorough analysis of partial participation; approximate aggregation to optimize communication efficiency; experimental evaluations on the integration of DP exploiting our system's scalability to compress peer-sampling rates.

\bibliography{main}
\bibliographystyle{abbrvnat}

\clearpage
\appendix
\section*{Appendix}
\label{sec:appendix}

\input{chapters/09_appendix}

\end{document}

%% file: chapters/01_introduction.tex
The convergence of Artificial Intelligence (AI) and next-generation wireless networks is driving a fundamental transformation in how we approach distributed computing and collaborative learning. As 6G and WiFi 9 standardization efforts begin to shape the future of global communication infrastructure, the ability to leverage distributed computational resources across wireless networks becomes not just an opportunity but a necessity for realizing the vision of AI-native wireless systems. The rapid proliferation of data across distributed sources -- from edge devices to base stations -- has created unprecedented opportunities for Machine Learning (ML), yet accessing and utilizing these dispersed data repositories remains a fundamental challenge~\citep{kairouz21,li20}. While centralized ML has driven remarkable advances in AI, it faces increasing limitations: data privacy regulations restrict data movement across organizational and geographical boundaries, bandwidth constraints in wireless environments make centralized data aggregation impractical, and the concentration of computational resources in large-scale data centers creates geographical and economic disparities in AI development capabilities~\citep{kairouz21,zhang20}. Federated Learning (FL) has emerged as a useful paradigm that enables collaborative model training over wide-area networks while keeping data localized, effectively tapping into vast data silos that would otherwise remain inaccessible for AI development~\citep{mcmahan17,kairouz21,li20,zhang20}. 

The promise of FL extends beyond privacy preservation to address a critical infrastructure challenge particularly relevant to next-generation wireless networks: the democratization of AI training capabilities at the network edge. Current AI development is increasingly dominated by regions with access to massive, centralized computing infrastructure and abundant power resources. However, many regions -- particularly in Europe -- face significant constraints in building comparable large-scale AI data centers due to power grid limitations, environmental regulations, and infrastructure costs~\citep{europaCapacitiesCrosszonal}. This disparity threatens to create a widening gap in AI capabilities between regions with different infrastructure capacities.

In the context of emerging wireless systems, where edge intelligence and distributed processing are fundamental design principles, FL offers a compelling alternative by enabling the orchestration of scattered computational resources -- from mobile devices to small cell base stations -- into a collective training infrastructure without requiring massive capital investments or power concentration~\citep{mcmahan17,kairouz21,li20}.

\begin{table}[t]
    \centering
    \caption{Related work overview. Our system is the first to deliver communication-efficient end-to-end P2P FL.}
    \label{tab:related_work}
    \resizebox{\linewidth}{!}{
    \begin{tabular}{@{}lcccccl}
    \toprule
    \thead{Approach} & \thead{Allows partial\\ communication} & \thead{Provides\\ global aggregation} & \thead{No sparsification} & \thead{Peer dropout\\ tolerance} & \thead{Enables private\\ training} & \thead{Source} \\
    \midrule
    RDFL         & --          & \checkmark     & \checkmark    & --            & -- & \citet{hu20}   \\
    SAPS         & \checkmark  & --             & --            & --            & -- & \citet{tang20} \\
    BrainTorrent & \checkmark  & --             & \checkmark    & \checkmark    & -- & \citet{roy19}  \\
    \hline
    \addlinespace[3pt]
    \textbf{MAR-FL (ours)} & \checkmark & \checkmark & \checkmark & \checkmark & \checkmark & \textit{This paper} \\
    \bottomrule
    \end{tabular}}
\end{table}

Peer-to-peer (P2P) FL systems represent the natural evolution of this distributed paradigm, aligning perfectly with the vision of AI-native wireless networks where intelligence is embedded throughout the network and does not require a centralized coordination server. By eliminating the central coordinator, P2P FL removes the communication and memory bottleneck of client-server FL -- where the server must coordinate massive numbers of unreliable devices in cross-device settings and shuttle large models in cross-silo settings -- thereby throttling scalability and slowing training~\citep{alqahtani19,huang23}. It also eliminates the single point of failure: progress no longer hinges on server-side compute or networking capacity, which can otherwise jeopardize training~\citep{tang20}. Freed from these constraints, P2P FL can harness available computational resources wherever they exist—from idle GPUs in edge servers to distributed computing nodes in radio access networks—creating a resilient, fault-tolerant training infrastructure that adapts to the dynamic resource availability inherent in wireless environments. This decentralized approach is particularly valuable in scenarios where network topology changes rapidly, devices join and leave unpredictably, and no single entity can or should control the training process (e.g., multi-operator collaborations or community-driven deployments).
These challenges create a fundamental research question: 
\emph{Can we design a communication-efficient P2P FL system that maintains training quality while handling the high peer churn rates and sudden training dropouts characteristic of wireless environments?}

\textbf{Contributions.}
In this paper, we present \textbf{Moshpit All-Reduce FL (MAR-FL)}, a novel P2P FL system that builds on dynamic iterative group formation to significantly improve communication efficiency and tolerance towards unexpected peer churn. 
MAR-FL allows scalable decentralized learning by reducing the overall communication load and the required number of interactions between peers. 
Our system incorporates Knowledge Distillation (KD) to boost training performance and supports optional Differential Privacy (DP) to mitigate remaining risks of private information leakage.
We conduct a comprehensive experimental evaluation that compares MAR-FL against client–server FL and P2P FL techniques, assessing communication efficiency, scalability, and robustness to network churn.

% In this paper, we present \textbf{Moshpit All-Reduce FL (MAR-FL)}, a novel P2P FL system that builds on dynamic iterative group formation to significantly improve communication efficiency and tolerance towards unexpected peer churn. 
% MAR-FL allows scalable decentralized learning by reducing the overall communication load and the required number of interactions between peers. 
% Our system incorporates Knowledge Distillation (KD) to boost training performance and leverages fully decentralized Differential Privacy (DP) to mitigate remaining risks of private information leakage. 
% We conduct a comprehensive experimental evaluation that compares MAR-FL against client–server FL and P2P FL methods, assessing communication efficiency, scalability, and robustness to network churn.

\textbf{Related work.}
Despite compelling advantages over client-server FL, existing P2P FL systems face severe practical limitations preventing deployment in bandwidth-limited wireless networks (\Cref{tab:related_work}). 
The Galaxy Federated Learning system's Ring Decentralized FL (RDFL)~\citep{hu20} incurs communication costs orders of magnitude higher than centralized FedAvg, making it economically infeasible for wireless environments. 
Moreover, RDFL's closed ring topology cannot tolerate the dynamic participation and node failures characteristic of wireless networks due to mobility, channel fading, or varying signal conditions. 
Sparsification and Adaptive Peer Selection (SAPS)~\citep{tang20} improves communication efficiency through model sparsification and single high-throughput peer exchanges per round, but spreads information only locally without synchronized global aggregation, slowing convergence and making progress sensitive to churn. 
BrainTorrent~\citep{roy19} provides serverless P2P flexibility through dynamic model fetching and merging, but relies on uncoordinated gossip-based learning that suffers from inefficient global information propagation and vulnerability to node churn.

\textbf{Structure.}
We introduce our new MAR-FL system in \Cref{sec:marfl} and evaluate it in \Cref{sec:results}. We conclude in \Cref{sec:conclusions}. % discuss broader implications and remaining hurdles in \Cref{sec:discussion} and 

%% file: chapters/02_marfl.tex
The overall objective of our P2P FL system is to reduce the communicational effort required to obtain globally averaged models, while retaining resilience to real-world network churn. Consequently, we deploy Moshpit All-Reduce (MAR) as fully decentralized aggregation mechanism.

\subsection{Problem Formulation}\label{subsection:marfl_problem}

We consider a P2P FL setting with $N$ peers, each holding a private local dataset $\mathcal{D}_i$, which may be heterogeneous and non-i.i.d. across peers. 
Training proceeds over $T$ iterations; in each iteration, peers perform local updates and exchange models over bandwidth-limited wireless links to conduct global aggregation. 
The system thereby faces the central FL challenge of communication costs: due to wireless links and connections operating at lower rates than intra- or inter-datacenter links, communication is costly and often by orders of magnitude slower than local computation~\citep{kairouz21,li20}. 
Consequently, our objective is to minimize the communication cost of P2P FL systems.

\subsection{Proposed System}\label{subsection:marfl_framework}

% \begin{wrapfigure}[15]{R}{0.48\textwidth}
%     \vspace{-12pt}
%     \begin{minipage}{\linewidth}
%         \begin{algorithm2e}[H]
%             \caption{MAR-FL (for $i$-th peer)}
%             \label{alg:marfl}
%             \scriptsize % otherwise \footnotesize
%             \KwInput{$\theta^0$, $m^0$, $\eta$, $\mu$, $D_i$, $B$, $T$, $N$, $\mathrm{use}_{\mathrm{kd}}$}
%             $(\theta_i^0, m_i^0)\leftarrow (\theta^0, m^0)$\\
%             \For{$t=1,2,\ldots,T$}{
%               \uIf{$i \in \mathcal{U}_t$}{
%                 $(\theta_i^t,m_i^t)\leftarrow \texttt{Momentum-SGD}(\theta_i^{t-1},m_i^{t-1},D_i,B,\eta,\mu)$
%               }\Else{
%                 $(\theta_i^t,m_i^t)\leftarrow (\theta_i^{t-1},m_i^{t-1})$
%               }
%               \If{$i \in \mathcal{A}_t$}{
%                 \If{$\mathrm{use}_{\mathrm{kd}}$}{
%                   $(\theta_i^t,m_i^t)\leftarrow \texttt{Moshpit-KD}(\theta_i^t;m_i^t,\mathcal{A}_t)$
%                 }
%                 $(\theta_i^t,m_i^t)\leftarrow \texttt{Moshpit-AR}(\theta_i^t;m_i^t,\mathcal{A}_t)$
%               }
%             }
%             \Return{$\theta_i^T$}
%         \end{algorithm2e}
%     \end{minipage}
%     \vspace{-6pt}
% \end{wrapfigure}

\begin{wrapfigure}[14]{R}{0.48\textwidth}
    \vspace{-14pt}
    \begin{minipage}{\linewidth}
        \begin{algorithm2e}[H]
            \caption{MAR-FL (for $i$-th peer)}
            \label{alg:marfl}
            \scriptsize % otherwise \footnotesize
            \KwInput{$\theta^0$, $m^0$, $\eta$, $\mu$, $D_i$, $B$, $T$, $N$, $\mathrm{use}_{\mathrm{kd}}$}
            \For{$t=1,2,\ldots,T$}{
              \uIf{$i \in \mathcal{U}_t$}{
                $(\theta_i^t,m_i^t)\leftarrow \texttt{Momentum-SGD}(\theta^{t-1},m^{t-1},D_i,B,\eta,\mu)$
              }\Else{
                $(\theta_i^t,m_i^t)\leftarrow (\theta^{t-1},m^{t-1})$
              }
              \If{$i \in \mathcal{A}_t$}{
                $\mathcal{S}_t := \{\, (j,\theta_j^{t},m_j^{t}) \mid j \in \mathcal{A}_t\}$\\
                \If{$\mathrm{use}_{\mathrm{kd}}$}{
                  $(\theta_i^t,m_i^t)\leftarrow \texttt{Moshpit-KD}(\mathcal{S}_t)$
                }
                $(\theta^t,m^t)\leftarrow \texttt{Moshpit-AR}(\mathcal{S}_t)$
              }
            }
            \Return{$\theta^T$}
        \end{algorithm2e}
    \end{minipage}
    \vspace{-6pt}
\end{wrapfigure}

\textbf{Integrating MAR into FL.}
For global model aggregation in fully decentralized FL, we adopt the idea of Moshpit SGD~\citep{ryabinin21}, where peers conduct MAR to dynamically form small independent groups and repeat this group matchmaking across multiple rounds until local information from all peers has propagated through the network. 
This procedure has two main benefits: global model averaging can be achieved without all-to-all communication, and peer dropouts only affect a single group (i.e., a very restricted number of peers). 
\fm{The overall MAR-FL training process (\Cref{alg:marfl}) starts for every peer $i\in[N]$ with the same randomly initialized model $\theta^0$ and momentum vector $m^0$, where $N$ denotes the total number of peers and $T$ the total number of FL iterations.
In each FL iteration $t\in\{1,\ldots,T\}$, every participating peer $i \in \mathcal{U}_t$, where $\mathcal{U}_t \subseteq [N]$, performs a local Momentum-SGD update on $B$ mini-batches of its local data $D_i$, using stepsize $\eta$ and momentum $\mu$. 
The update follows the damped momentum update proposed by~\citet{reddi20} and yields a local peer state $(\theta_i^t,m_i^t)$. 
A set of aggregation peers $\mathcal{A}_t\subseteq[N]$ then performs MAR on its aggregation state set $\mathcal{S}_t$, where
$\mathcal{S}_t := \{ (j,\theta_j^{t},m_j^{t}) \mid j \in \mathcal{A}_t \}$, to obtain a globally averaged state $(\theta^t,m^t)$. 
\cameradone{This is done in multiple group formation rounds $g \in G^{t}$ per FL iteration $t$.}
KD is integrated if the $\mathrm{use}_{\mathrm{kd}}$ flag is set. 
After $T$ iterations, each peer holds the final collaboratively trained global model $\theta^T$.}

\textbf{Coordinating FL peers.}
Synchronization of peers during group formation is coordinated through Distributed Hash Tables (DHT).
\cameradone{Our system thereby relies on a Hivemind Kademlia DHT solely for lightweight coordination -- barriers and group-formation metadata -- while model and momentum weights never traverse the DHT. 
A single DHT get/store involves at most $\mathcal{O}(\log{N})$ hops. 
In our implementation, coordination occasionally scans peer announcements (issuing $\mathcal{O}(N)$ look-ups), so the control-plane cost per round is $\mathcal{O}(N\log{N})$ and remains negligible compared to model-exchange traffic.}
To assemble into groups, each peer manages its own group key and forms groups with peers sharing the same key value in the DHT. 
To avoid redundant information exchange in consecutive MAR rounds, peers are prevented from revisiting one another within a single FL iteration by group key initialization and updates that leverage their chunk indices from $d-1$ previous MAR rounds. 
We therefore adopt techniques proposed by~\citet{ryabinin21}. 
\cameradone{In an optimal MAR-FL setup, exact global averaging can be achieved after $d$ rounds of MAR when the group size is $M$ and the group key dimension is $d$, so that the total number of peers $N$ satisfies $N = M^d$. 
With fixed MAR group size $M$, each round makes a peer talk to at most $(M-1)$ others, and achieving (near-)global averaging needs $G \approx \lceil \log_{M} N \rceil$ rounds (exactly $G{=}d$ when $N{=}M^{d}$). 
Hence, each peer performs $\mathcal{O}(\log_{M} N)$ exchanges per iteration and, over all peers, the system incurs $\mathcal{O}(N \log_{M} N)=\mathcal{O}(N\log N)$ communication per iteration, versus $\mathcal{O}(N^{2})$ for P2P FL systems using all-to-all communication.}

\begin{wrapfigure}[21]{R}{0.50\textwidth}%0.5
    \vspace{-14pt}
    \begin{minipage}{\linewidth}
        \begin{algorithm2e}[H]
            \caption{Moshpit-{KD} (for $i$-th peer in MKD round $g$ of FL iteration $t$)}
            \label{alg:mkd}
            \scriptsize
            \KwInput{$\theta_i^{g-1}$, $m_i^{g-1}$, $\mathcal{C}_g$, $\{\theta_c^{\,g-1}\}_{c \in C_g}$, $\mathcal{B}$, $E$, $\eta$, $\mu$, $\tau$, $\rho_\ell$, $K$}
            \KwOutput{$\theta_i^{g}$, $m_i^{g}$}
            $(\mathcal{C}_g^{\text{top}},\, \ell,\, \{\,z_b^{(c)}\,\}_{b\in\mathcal{B},\, c \in \mathcal{C}_g^{\text{top}}}\big) \leftarrow \texttt{TeacherSel}(\theta_i^{g-1}, m_i^{g-1}, \mathcal{C}_g, \{\theta_c^{\,g-1}\}_{c \in C_g}, \mathcal{B}, \tau, \rho_\ell)$\\
            \For{$b\in\mathcal{B}$}{
                $\bar z_b \leftarrow \frac{1}{\ell}\!\sum_{c\in\mathcal{C}_g^{\text{top}}} z_b^{(c)}$.
            }
            $(\theta_i^{g}, m_i^{g}) \leftarrow (\theta_i^{g-1}, m_i^{g-1})$\\
            \For{$e=1,2,\ldots,E$}{
                \For{$b \in \mathcal{B}$}{
                    $\,\hat s_b \gets f_{\theta_i^{g}}(x_b)$\\
                    $L_{\mathrm{KL}} \gets \tau^2 \cdot D_{\mathrm{KL}}\!\big(\mathrm{softmax}(\bar z_b/\tau)\,\big\|\,\mathrm{softmax}(\hat s_b/\tau)\big)$\\
                    $L_{\mathrm{CE}} \gets \mathrm{CE}\big(y_b,\mathrm{softmax}(\hat s_b)\big)$\\
                    $\lambda \gets \max\!\big(0,\,1-\tfrac{t-1}{K}\big)$\\
                    $L \gets \lambda\,L_{\mathrm{KL}} + (1-\lambda)\,L_{\mathrm{CE}}$\\
                    $m_i^{g} \gets \mu\, m_i^{g} + (1-\mu)\,\nabla_{\theta_i^{g}} L$\\
                    $\theta_i^{g} \gets \theta_i^{g} - \eta\, m_i^{g}$\\
                }
            }
            \Return $\theta_i^{g}, m_i^{g}$\\
        \end{algorithm2e}
    \end{minipage}
\end{wrapfigure}

\textbf{Concept of KD.}
Our MAR-FL system allows the integration of KD to accelerate model convergence. 
\fm{
Let $C_g \subseteq \mathcal{A}_t$ be the candidate teacher peers in MKD round $g$ with local models $\{\theta_c^{\,g-1}\}_{c\in C_g}$, where $\mathcal{A}_t$ refers to~\Cref{alg:marfl}. Candidate teachers are collected using the same procedure MAR uses for global model averaging; hence, we call this mechanism Moshpit-KD (MKD).} 
The MKD process of an entire FL iteration proceeds over multiple MKD rounds $g \in \{1,\ldots,G\}$, where each round $g$ includes group formation and candidate teacher collection followed by the actual distilling of knowledge. 
To balance model utility and communication overhead, we use MKD only in the first $K$ FL iterations. 
\Cref{alg:mkd} illustrates MKD round $g$ in FL iteration $t \in \{1,\ldots,K\}$, where $K \leq T$ denotes the number of FL iterations in which we actually apply MKD, with $T$ being the total number of MAR-FL iterations in~\Cref{alg:marfl}. 
To account for data heterogeneity in FL~\citep{shao24}, MKD selects a subset of top-$\ell$ teachers $\mathcal{C}_g^{\text{top}} \subseteq \mathcal{C}_g$ with the lowest Kullback–Leibler (KL) divergence, where $\rho_\ell$ is the selection ratio (details in~\Cref{subsection:app_mkd}). 
Student $i$ then distills knowledge from these selected teachers: over $E$ local epochs, starting from the previous MKD round's state $(\theta_i^{g-1},m_i^{g-1})$, the student updates on each available local mini-batch $b\!\in\!\mathcal{B}$ by computing a student loss $L$ and applying Momentum-SGD~\citep{reddi20} with learning rate $\eta$ and momentum $\mu$ to eventually obtain an updated state $(\theta_i^{g}, m_i^{g})$. 
In MKD round $g=1$, the previous state $(\theta_i^{0},m_i^{0})$ refers to the student's state before any MKD is applied (i.e., after local model update).
The student loss $L$ aligns to the loss term proposed by~\citet{hinton15}: $L$ is the weighted sum of the KL divergence $D_{\mathrm{KL}}$ between softened probability distributions over teacher-ensemble and student classes, rescaled by the squared temperature $\tau^2$, and a CE term $L_{\mathrm{CE}}$ on hard labels $y_b$. 
Averaged teacher-ensemble logits are hereby denoted as $\bar z_b$ and student logits as $\hat s_b$. 
As we use MKD only in the first $K$ FL iterations, we facilitate a gradual transition from the use of MKD to its complete omission by linearly reducing the weighting $\lambda$ of the KL term $L_{KL}$.

\textbf{Privacy considerations.}
To allow privacy preserving training, we adapt the DP-FedAvg with adaptive clipping~\citep{andrew21} to fit our serverless P2P system (\Cref{algo:dpmarfedavg}, see~\Cref{subsection:app_fddp}). 
In each FL iteration, every peer first computes the difference between its current local model and the previously aggregated global model.
This update is then clipped to an adaptive bound and perturbed with Gaussian noise. 
The privatized update is used to compute a DP-safe local model and peers run MAR. 
After the final round of MAR, the clipping bound is updated to track a globally averaged clipping rate. 
This procedure fully decentralizes DP with adaptive clipping and renders it ready to use with MAR-FL: privacy loss accrues entirely from local computations, while MAR merely averages privatized models across groups.

\subsection{Convergence Analysis}

The convergence of MAR-FL follows from the model mixing dynamics of MAR, analyzed by~\citet{ryabinin21}. 
In the optimal case where the total number of peers $N$ forms a perfect $d$-dimensional grid $N=M^d$ and there are no peer dropouts, MAR computes the exact global average after exactly $d$ rounds of communication -- i.e., within a single FL iteration $t$ when that iteration schedules $d$ MAR rounds. 
For general settings, MAR exhibits exponential convergence to the global average $\bar \theta$. 
Specifically, if peers are randomly partitioned each iteration into $r$ groups that average locally, the expected average distortion after $T$ averaging iterations satisfies:

\begin{equation}
\mathbb{E}\left[ \frac{1}{N} \sum_{i=1}^{N} \|\theta_i^T - \bar \theta\|^2 \right] = \left( \frac{r - 1}{N} + \frac{r}{N^2} \right)^T \frac{1}{N} \sum_{i=1}^{N} \|\theta_i - \bar \theta\|^2.
\end{equation}

While this rate is derived for a simplified random-grouping model, our system's MAR implementation avoids revisiting peers via deterministic key updates, which in practice accelerates mixing relative to purely random grouping. 
Crucially, the bound is independent of the spectral properties of the communication graph, avoiding the scaling limits typical of gossip-based decentralized FL.

%% file: chapters/03_experiments.tex
\begin{wrapfigure}[38]{R}{0.40\textwidth}
  \vspace{-32pt}%-12
  \centering
  % ---- EF ----
  \begin{minipage}[t]{\linewidth}
    \centering
    \captionsetup[subfigure]{justification=centering}
    \includegraphics[width=\linewidth]{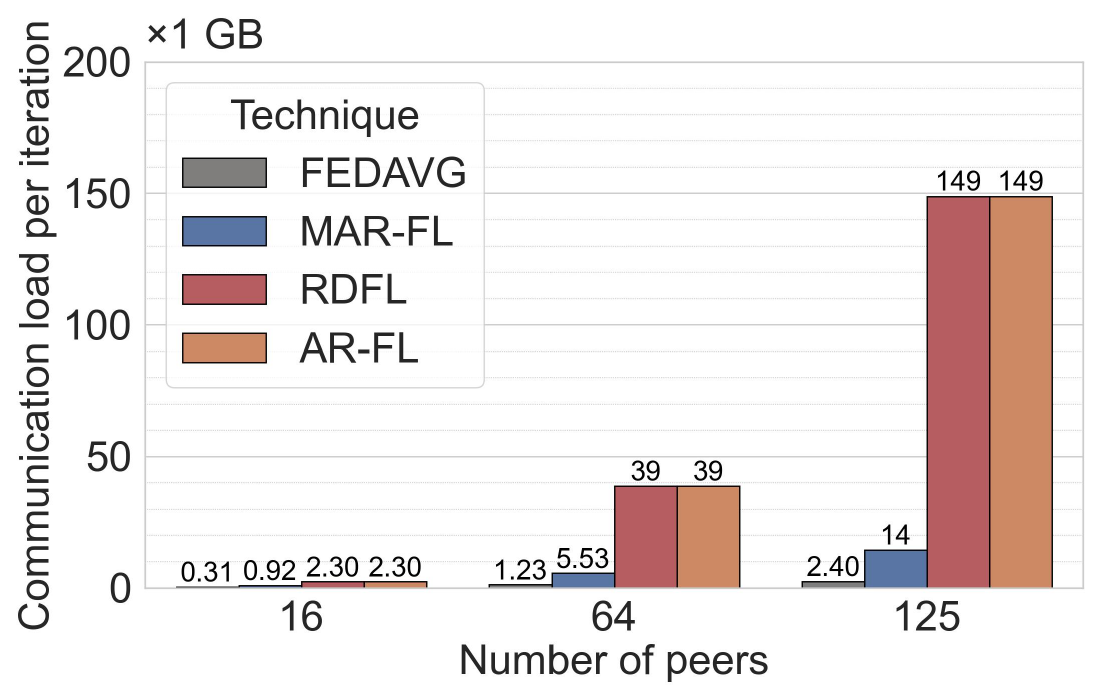}
    \subcaption{MNIST}
  \end{minipage}\hfill
  \begin{minipage}[t]{\linewidth}
    \centering
    \captionsetup[subfigure]{justification=centering}
    \includegraphics[width=\linewidth]{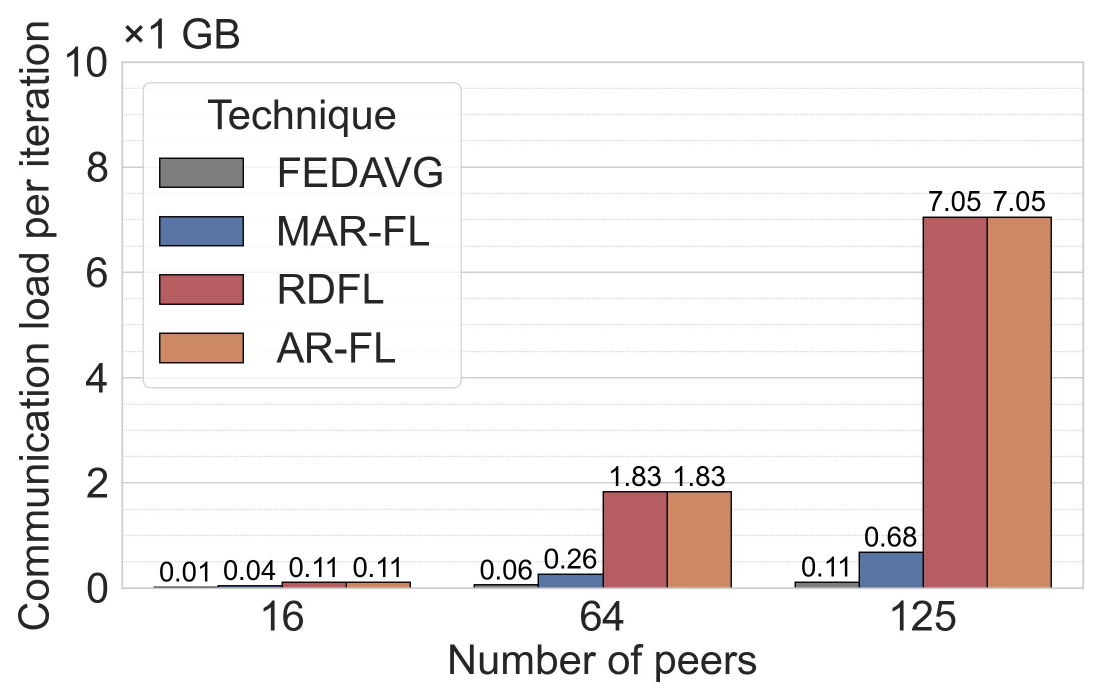}
    \subcaption{20NG}
  \end{minipage}
  \captionsetup{skip=5pt}%5
  \caption{\cameradone{Performance gap evaluation.} MAR-FL improves communication efficiency by up to $10\times$ compared to existing P2P FL baselines.}
  \label{fig:e12_comm_scalability}
  \vspace{8pt}%8
  % ---- KD ----
  \begin{minipage}[t]{\linewidth}
    \centering
    \includegraphics[width=\linewidth]{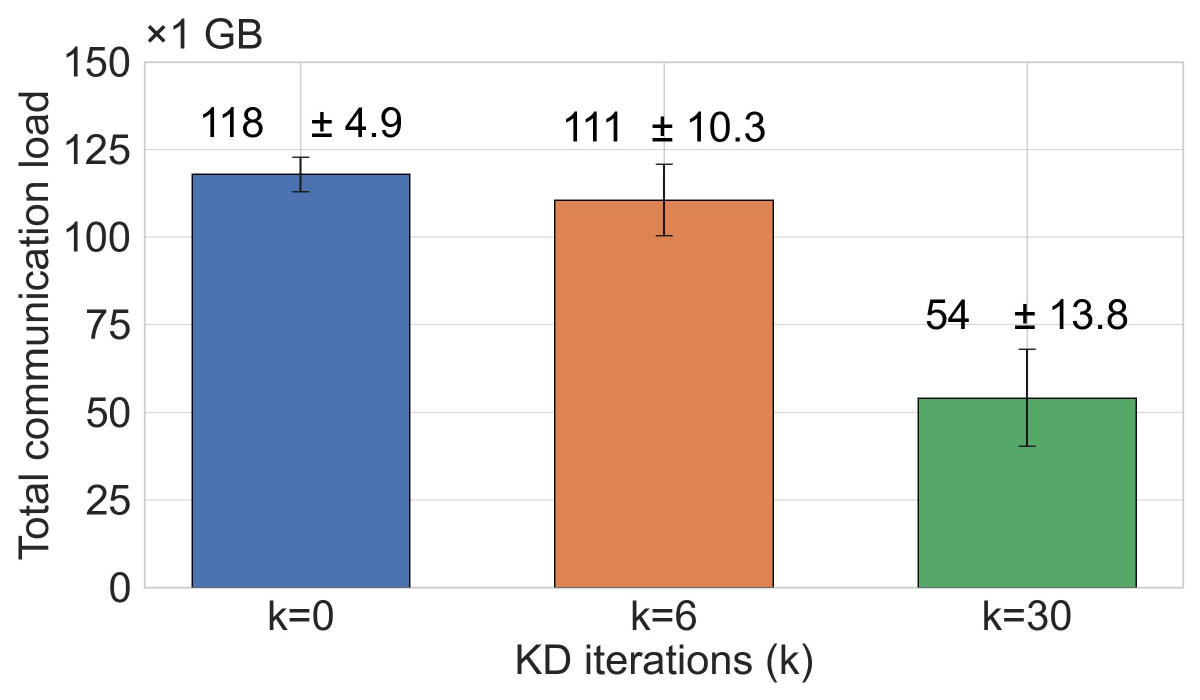}
  \end{minipage}
  \captionsetup{skip=5pt}%5
  \caption{With MKD, the communication load of systems using MAR-FL is further reduced, as we require over $2\times$ less communication to reach $50\%$ accuracy. Plot shows results on 20NG. Results on MNIST are available in the appendix.}
  \label{fig:e22_kd_20ng}
\end{wrapfigure}

This section presents our experimental setup and evaluates results in detail, while emphasizing communication cost, scalability, robustness and trade-offs concerning model utility. 

\subsection{Setup}\label{subsection:exp_setup}
In the following we delineate ML datasets and models, reference baselines, and parametrization used to evaluate and contextualize MAR-FL. 
Underlying objectives are described. 
We use a simulation environment for all of our experiments. 
Due to constraints of our simulation environment, model evaluation is conducted every fifth FL iteration. 
We simulate all experiments on a single node with 4$\times$H100 GPUs, 768\,GB of memory, and 96 CPU cores.
Our code is publicly available.\footnote{Github: \url{https://github.com/felix-fjm/mar-fl}}
Additional details on the experimental setup can be found in the appendix.

\textbf{Datasets and models.}
\fm{We evaluate MAR-FL on two widely used ML datasets,} namely MNIST~\citep{lecun2010mnist} and 20 Newsgroups (20NG)~\citep{LANG1995331}. 
For MNIST, we use a CNN-based architecture and for 20NG we use a frozen DistilBERT model~\citep{Sanh19} with a classification head. 
We employ a Latent Dirichlet Allocation ($\alpha = 1.0$) to create non-i.i.d. subsets for 16, 64, and 125 FL peers. 
If not specified otherwise, experiments run on 125 FL peers.
Per aggregation round, each peer trains on 64 and 16 samples for MNIST and 20NG, respectively.

\begin{wrapfigure}[36]{R}{0.40\textwidth}
  \centering
  % ---- PP ----
  \begin{minipage}[t]{\linewidth}
    \centering
    \captionsetup[subfigure]{justification=centering}
    \includegraphics[width=\linewidth]{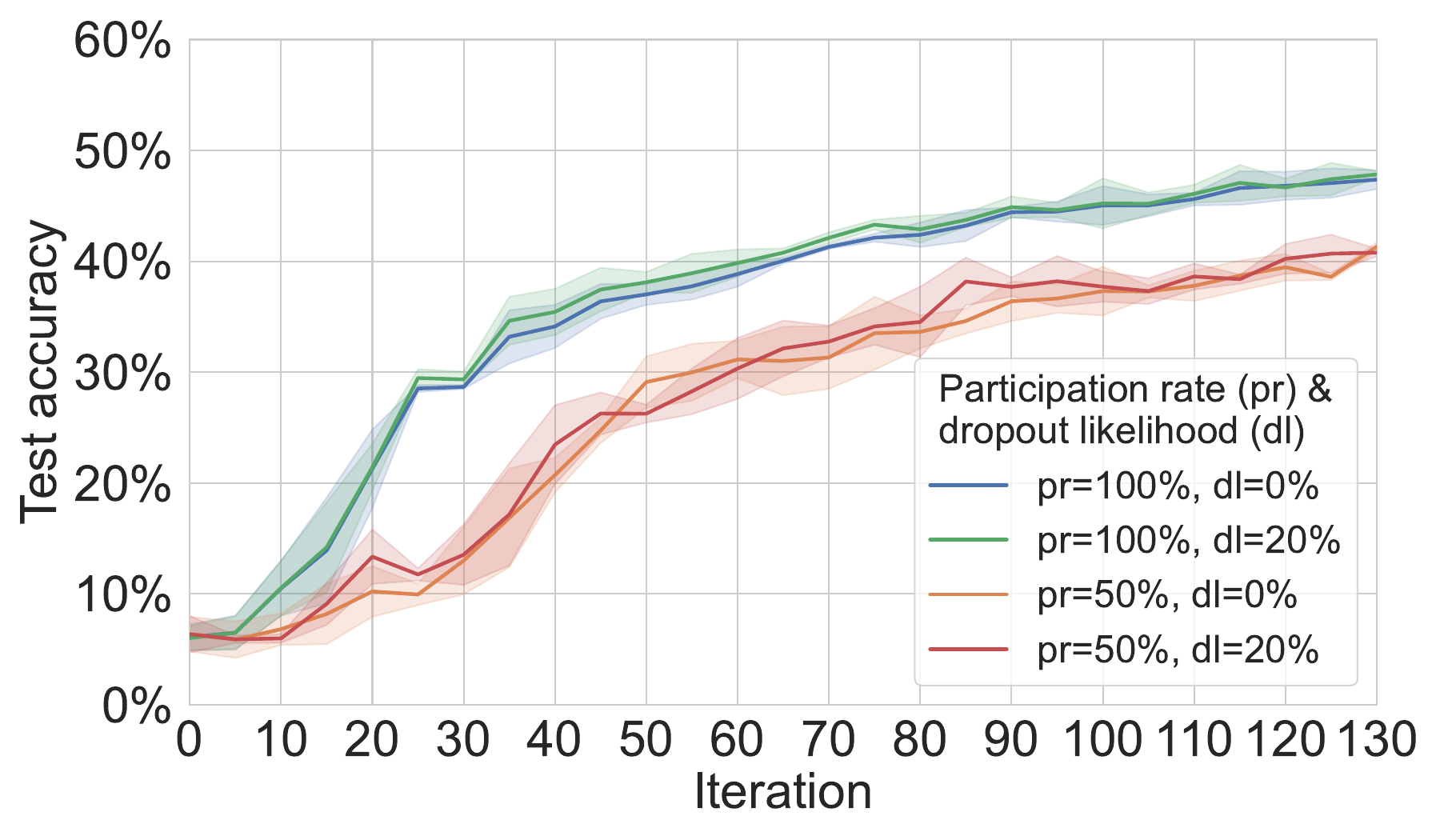}
    \subcaption{Model Performance -- MAR-FL -- 20NG}
  \end{minipage}\hfill
  \begin{minipage}[t]{\linewidth}
    \centering
    \captionsetup[subfigure]{justification=centering}
    \includegraphics[width=\linewidth]{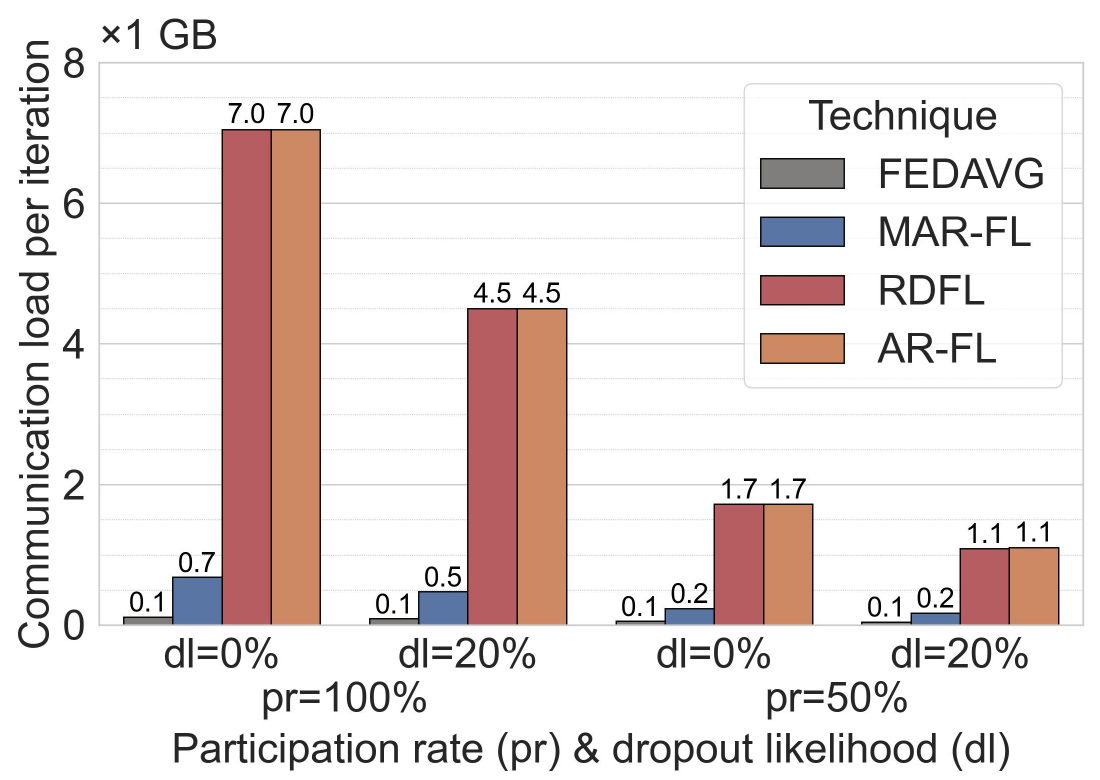}
    \subcaption{Communication Cost -- 20NG}
  \end{minipage}
  \captionsetup{skip=5pt}%5
  \caption{MAR-FL is affected by partial participation but resilient towards sudden dropouts.}
  \label{fig:e4_partial_participation_news}
  \vspace{8pt}%8
  % ---- DP ----
  \begin{minipage}[t]{\linewidth}
    \centering
    \includegraphics[width=\linewidth]{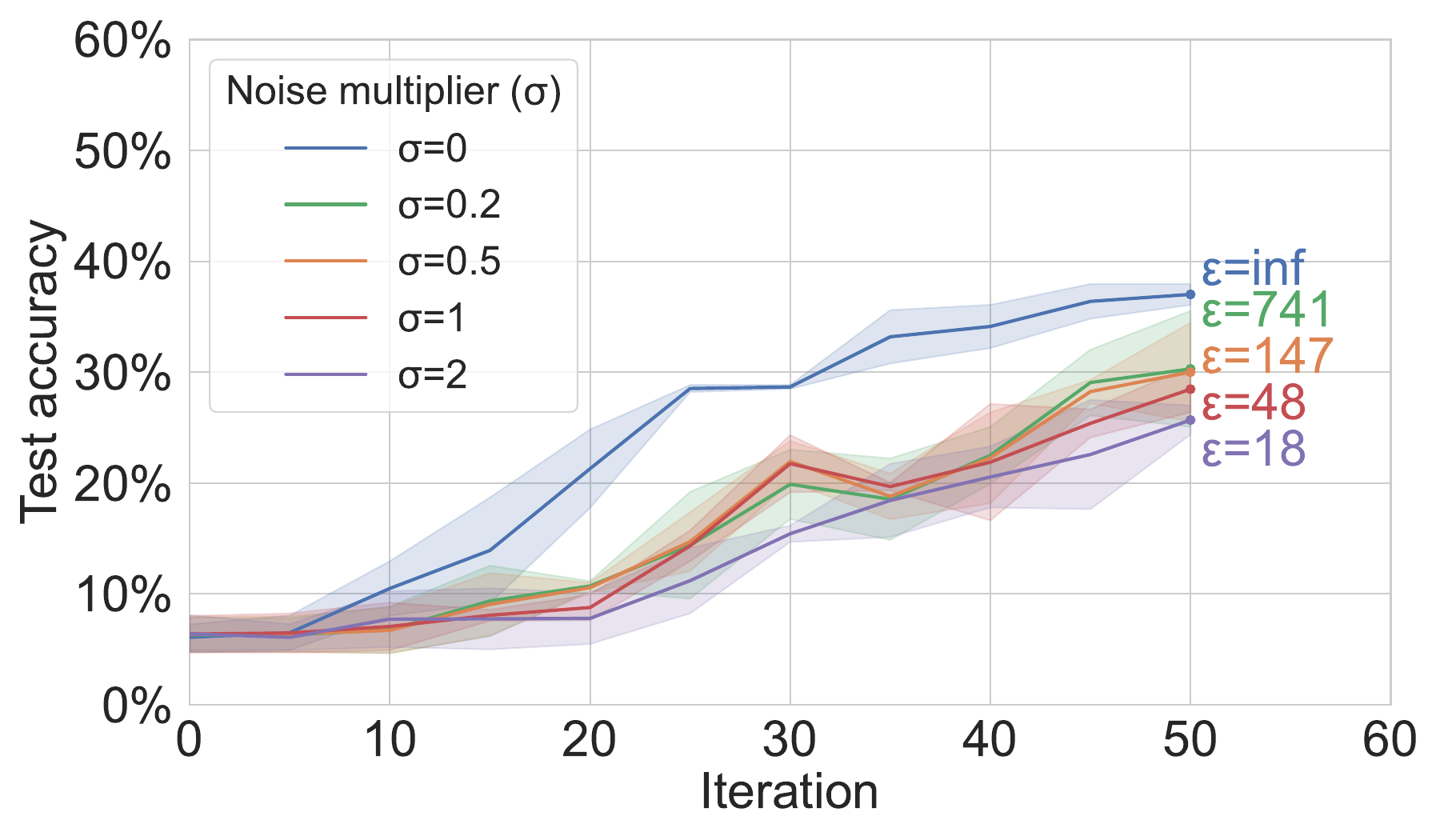}
  \end{minipage}
  \captionsetup{skip=5pt}%5
  \caption{MAR-FL is compatible with DP and exhibits the same performance characteristics as FedAvg. Plot shows results on 20NG. Results on MNIST are available in the appendix.}
  \label{fig:e31_dp_20ng}
\end{wrapfigure}

\textbf{FL baselines.}
We directly compare MAR-FL to the client-server FedAvg standard and established P2P FL techniques, specifically RDFL (ring all-reduce), which is at the center of the Galaxy Federated Learning framework.\footnote{We do not compare against Galaxy Federated Learning as a whole since the framework largely depends on a distributed ledger/blockchain for training verification. Verification is beyond the scope of our work.} 
We further evaluate MAR-FL against a naïve all-to-all All-Reduce FL algorithm (AR-FL), where all peers communicate with each other.
\cameradone{Even though we discuss BrainTorrent and SAPS in our related work section, their limitations regarding communication efficiency make practical deployments prohibitively expensive, which is why we omit the two techniques as baselines.}

\textbf{Parametrization of MAR-FL.}
We use exact aggregation of peer groups, if not specified otherwise.
For evaluating the compound benefits of MAR-FL and KD, we use a teacher selection ratio $\rho_\ell=0.4$~\citep{hu20}, student loss temperature $\tau=3.0$~\citep{hinton15} and one epoch. Parameter choices for adaptive DP align to~\citet{andrew21} and are listed in~\Cref{subsection:app_fddp}.

\textbf{Local model aggregation.} 
For peer-side local aggregation, we use SGD with momentum~\citep{reddi20}, set the learning rate to $\eta = 0.1$ and the momentum to $\mu = 0.9$. 
Across techniques, we use full peer participation if not specified otherwise (typical setup for cross-silo FL applications).

\textbf{Partial participation and network churn.}
To assess the effect of partial participation and network churn, we vary participation rates and dropout likelihoods. 
Participation rates control how many peers participate in an entire FL iteration consisting of local updates and global aggregation, while dropout likelihoods simulate unreliable peer connectivity (i.e., peer has conducted local update but does not participate in global aggregation). 

\textbf{Privacy.}
To investigate privacy-preserving training and its effect on model utility, we vary the noise multiplier to control the extent of privatization. 
The peer sampling rate, where lower values reduce the privacy loss, is fixed at $100\%$.
Results on scalability and partial participation will reveal whether our system can leverage this rate to enhance privacy without degrading training performance.

\subsection{Results}\label{subsection:exp_results}

\textbf{Communication efficiency and scalability.}
Across both ML tasks (MNIST and 20NG), MAR-FL matches the training performance of all three baselines (see~\Cref{subsection:app_qrmarflbaselines}). 
This parity is expected because, with suitable MAR parameters, each iteration of MAR-FL attains an exact global average (e.g., group size $5$ and $3$ MAR rounds for $125$ peers: $125=5^3$). 
While obtaining identical model utility, MAR-FL requires far less communication per iteration, up to $10\times$ less communication than RDFL or AR-FL. The communication complexity of MAR-FL, $\mathcal{O}(N\log{N})$, yields stronger performance as systems scale (\Cref{fig:e12_comm_scalability}).
In contrast, RDFL and AR-FL exhibit a complexity of $\mathcal{O}(N^2)$.

\textbf{Improving communication efficiency with MKD.}
MAR-FL achieves substantially higher communication efficiency than our P2P FL baselines, narrowing the gap to the client-server FedAvg standard. 
\fm{To improve communication efficiency even further, MKD can be used. 
MKD accelerates model convergence so that a target accuracy can be reached with less total communication (\Cref{fig:e22_kd_20ng}), although increasing the per-iteration load.} 
The trade-off between communication costs and model utility can be controlled by the number of KD iterations.

\textbf{Partial participation and network churn.}
Partial participation leads to a substantial degradation of MAR-FL's training performance, while configured network churn and unreliable connectivity do not cause additional accuracy drops (\Cref{fig:e4_partial_participation_news}); our three baselines show the same pattern. 
\fm{While the training performance of all three P2P techniques is equally affected by these real-world system disturbances, MAR-FL consistently preserves its \hw{net benefit over all baselines} in communication efficiency, providing evidence for the enhanced practicality of our system.}
\fm{Even with $50\%$ participation and $20\%$ dropout likelihood, RDFL and AR-FL require more than $5\times$ the communication of MAR-FL to reach the same model utility.}
The robustness towards unreliable connectivity (i.e., peer has conducted local update but does not participate in global aggregation) can be attributed to the fact that averaging incomplete global models over multiple FL iterations eventually converges to almost exact global averages. 
In~\Cref{subsection:app_qrmarfl}, where we provide further results on partial participation, we outline how this phenomenon can be exploited to increase the communication efficiency of MAR-FL.
\cameradone{The appendix also includes additional results for FedAvg, RDFL, and AR-FL.}

\textbf{Differentially private training.}
When conducting DP-safe model aggregation in MAR-FL, increasing the strength of DP by raising the noise multiplier $\sigma$ reduces the privacy loss $\varepsilon$ but eventually degrades model utility (\Cref{fig:e31_dp_20ng}). 
Since our observations align with the effect of DP on standard FedAvg~\citep{andrew21,wei20}, this confirms that DP is readily supported within our fully decentralized system. 
We emphasize that the privacy loss $\varepsilon$ can be substantially reduced by decreasing the peer-sampling rate (i.e., partial participation in local updates)~\citep{wei20,mironov17}, so that our communication-efficient and scalable MAR-FL system provides a foundation for comprehensive privacy preservation.

%% file: chapters/09_appendix.tex
\section{Additional Details on the MAR-FL Method}

\subsection{Details on MKD}\label{subsection:app_mkd}

\begin{wrapfigure}[15]{R}{0.6\textwidth}
    \vspace{-10pt}
    \begin{minipage}{\linewidth}
        \begin{algorithm2e}[H]
            \caption{Teacher selection in MKD (for $i$-th peer in MKD round $g$ of FL iteration $t$)}
            \label{alg:mkd_ts}
            \scriptsize
            \KwInput{$\theta_i^{g-1}$, $m_i^{g-1}$, $\mathcal{C}_g$, $\{\theta_c^{\,g-1}\}_{c \in C_g}$, $\mathcal{B}$, $\tau$, $\rho_\ell$}
            \KwOutput{$\mathcal{C}_g^{\text{top}}$, $\ell$, $\{\,z_b^{(c)}\,\}_{b\in\mathcal{B},\, c \in \mathcal{C}_g^{\text{top}}}$}
            \For{$b \in \mathcal{B}$}{
                $s_b \leftarrow f_{\theta_i^{g-1}}(x_b)$
            }
            \For{$c \in \mathcal{C}_g$}{
                \For{$b \in \mathcal{B}$}{
                    $z_b^{(c)} \leftarrow f_{\theta_{c}^{g-1}}(x_b)$
                }
                $\,S(c) \leftarrow \sum_{b\in\mathcal{B}} D_{\mathrm{KL}}\!\big(\mathrm{softmax}(z_b^{(c)}/\tau)\,\big\|\,\mathrm{softmax}(s_b/\tau)\big)$.
            }
            $\ell \leftarrow \max\{1,\lfloor \rho_\ell\,|\mathcal{C}_g| \rfloor\}$.\\
            $\mathcal{C}_g^{\text{top}} \leftarrow \ell\ \text{teachers with smallest}\ S(c)$.\\
            \Return $C_g^{\text{top}}$, $\ell$, $\{\,z_b^{(c)}\,\}_{b\in\mathcal{B},\, c \in C_g^{\text{top}}}$\\
        \end{algorithm2e}
    \end{minipage}
    \vspace{-6pt}
\end{wrapfigure}

\textbf{Teacher selection in MKD.} 
In MKD round $g$, candidate teachers $\mathcal{C}_g$ depict a collected subset of participating aggregation peers $\mathcal{A}_t$ of FL iteration $t$ from~\Cref{alg:marfl}. 
From these candidate teachers, a top-$\ell$ ratio is selected for actual distilling of knowledge. 
A peer $i$ thereby selects $\ell$ teachers $\mathcal{C}_g^{\text{top}} \subseteq \mathcal{C}_g$ which yield the $\rho_\ell$ smallest KL divergence when for each candidate teacher model $\{\theta_c^{\,g-1}\}_{c \in C_g}$ comparing the softened output distribution $\mathrm{softmax}(z_b^{(c)}/\tau)$ with its own softened output distribution $\mathrm{softmax}(s_b/\tau)$. 
The softening of output distributions is conducted by normalizing the logits $z_b^{(c)}$ and $s_b$ with a temperature $\tau$. 
Student logits $s_b$ are obtained by passing local mini-batches $\mathcal{B}$ through the student model $\theta_i^{g-1}$, while teacher logits $z_b^{(c)}$ are obtained by passing local mini-batches $\mathcal{B}$ through a candidate teacher model $\theta_c^{\,g-1}$. 
We use the KL-based rating of collected peer models to account for non-i.i.d. data in FL. 
\citet{shao24} emphasize that non-i.i.d. data distributions depict a crucial challenge for KD in FL, because local models cannot produce meaningful predictions on data outside of their own distributions. 
Even softening of output distributions is not solving this issue, as ensembles of inconsistent local predictions still exhibit high entropy, which leads to distilling ambiguous and misleading knowledge. 

% While~\citet{hu20} use teacher selection to avoid data poisoning from malicious peers, we use the underlying KL-based rating of collected peer models to account for non-i.i.d. data in FL. 

% While we align to the selection technique from~\citet{hu20}, they change the top-$\ell$ teacher ratio $\rho_\ell$ over time; however, due to our differing objectives, we keep $\rho_\ell$ constant.

\textbf{Deriving the MKD student loss.} 
In~\Cref{alg:mkd} we denote the student loss term $L$ as weighted sum of the KL divergence $D_{\mathrm{KL}}$ between softened probability distributions over teacher-ensemble and student classes, rescaled by the squared temperature $\tau^2$, and a CE term $L_{\mathrm{CE}}$ on hard labels $y_b$. 
This student loss can be derived from the student loss proposed by~\citet{hinton15}. 
Let $p_z = \mathrm{softmax}(z / \tau)$ be the teacher distribution and $p_s = \mathrm{softmax}(s / \tau)$ the student distribution at the same temperature $\tau$, where teacher logits are denoted by $z$ and student logits by $s$. 
Higher values of $\tau$ shrink differences between logits so that the distribution is softer, which can reveal relative similarities among classes (i.e., dark knowledge). 
\citet{hinton15} train the student with a weighted sum of CE to soft targets at $\tau > 1$ and CE to the hard labels $y$ at $\tau = 1$, scaling the soft-target gradients by $\tau^2$:
\begin{equation}
    L_\mathrm{Hinton} = (1-\alpha)\ \mathrm{CE}(y, \mathrm{softmax}(s)) + \alpha \tau^2\ \mathrm{CE}(p_z, p_s).
\end{equation}
Starting from this two-term objective, one can use the identity
\begin{equation}\label{equation:ceklequivalence}
    \mathrm{CE}(p_z, p_s) = H(p_z) + D_\mathrm{KL}(p_z \parallel p_s).
\end{equation}
and note that $H(p_z)$ is constant with respect to the student. Dropping that constant and absorbing $\alpha \tau^2$ into the KL weight gives
\begin{equation}
    L \equiv (1-\alpha)\ \mathrm{CE}(y, \mathrm{softmax}(s)) + \alpha \tau^2\, D_\mathrm{KL}(p_z \parallel p_s),
\end{equation}
which is the student loss term $L$ used in our MKD approach when $\alpha = \max\!\big(0,\,1-\tfrac{t-1}{K}\big)$.

\subsection{Fully Decentralized DP}\label{subsection:app_fddp}

\begin{wrapfigure}[25]{R}{0.58\textwidth}
    \vspace{-10pt}
    \begin{minipage}{\linewidth}
        \begin{algorithm2e}[H]
            \caption{DP-safe model aggregation in MAR-FL (for $i$-th peer in FL iteration $t$)}
            \label{algo:dpmarfedavg}
            \scriptsize
            \KwInput{$G$, $\mathcal{A}_t$, $n_t$, $\theta_i^{t}$, $m_i^{t}$, $\bar\theta_i^{t-1}$, $\bar{\Delta}_i^{t-1}$, $\beta$, $\sigma_\mathrm{mult}$, $C_t$, $\gamma$, $\eta_u$, $\eta_C$}
            \KwOutput{$\theta^{t}$, $m^{t}$, $C_{t+1}$, $\bar{\Delta}^{t}$}
            $\sigma_b \gets n_t/20$\\
            $z_\Delta \gets \left(\sigma_\mathrm{mult}^{-2}-(2\sigma_b)^{-2}\right)^{-\sfrac{1}{2}}$\\
            $\sigma_\Delta \gets z_\Delta C^t$\\
            $\Delta_i \gets \theta_i^{t} - \bar\theta_i^{t-1}$\\
            $b_i^{t,0} \;\gets\; \mathbf{1}\!\left\{\, \|\Delta_i\| \leq C_t \,\right\}$\\
            $\widetilde{\Delta}_i \gets \Delta_i \cdot \min\!\left(1,\frac{C_t}{\|\Delta_i\|}\right) + \mathcal{N}\!\left(0,\, I\frac{\sigma_\Delta^2}{n_t}\right)$\\
            $\bar{\Delta}_i^{t,0} \gets
            \begin{cases}
                \beta\,\bar{\Delta}_i^{t-1} + \widetilde{\Delta}_i, & \bar{\Delta}_i^{t-1} \neq \bot \\
                \widetilde{\Delta}_i, & \text{otherwise}
            \end{cases}$\\
            $\hat\theta_i^{t,0} \gets \bar\theta_i^{t-1} + \eta_u\,\bar{\Delta}_i^{t,0}$\\
            $m_i^{t,0} \gets m_i^{t}$\\
            \For{$g=1,2,\ldots,G$}{
                $\mathcal{P}_t := \{\, (j,\hat{\theta}_j^{t,g-1},\, m_j^{t,g-1},\, b_j^{t,g-1},\, \bar{\Delta}_j^{t,g-1}) \mid j \in \mathcal{A}_t \,\}$\\
                \uIf{$g<G$}{
                    $(\hat\theta_i^{t,g},\, m_i^{t,g},\, b_i^{t,g},\, \bar{\Delta}_i^{t,g}) \gets \texttt{MAR}_g(\mathcal{P}_t)$\\
                }\Else{
                    $(\theta^{t},\, m^{t},\, \bar{b}^{t},\, \bar{\Delta}^{t}) \gets \texttt{MAR}_g(\mathcal{P}_t)$\\
                    $\tilde{b}^{\,t} \gets \bar{b}^t + \frac{\mathcal{N}(0,\,\sigma_b^2)}{n_t}$\\
                    $C_{t+1} \gets C_t \cdot \exp\!\big(-\eta_C\,(\tilde{b}^{\,t}-\gamma)\big)$\\
                }
            }
            \Return $\theta^{t},\ m^{t},\ C_{t+1},\ \bar{\Delta}^{t}$\\
        \end{algorithm2e}
    \end{minipage}
    \vspace{-6pt}
\end{wrapfigure}

\citet{andrew21} propose DP-FedAvg with adaptive gradient clipping for client-server FL, in which a central server mediates the DP-safe model aggregation. 
We adapt this approach to fit our serverless P2P system. 
Our system's DP-safe model aggregation illustrated in~\Cref{algo:dpmarfedavg} corresponds to MAR in~\Cref{alg:marfl} when DP is activated. 
For simplicity, the local pre-aggregation state $(\theta_i^{t},\, m_i^{t})$, the peer’s last obtained global model $\bar\theta_i^{t-1}$, and the peer’s last obtained smoothed delta $\bar{\Delta}_i^{t-1}$ are all denoted as if peer $i$ had participated in the previous local update and aggregation. 
This is not necessarily the case, since our system allows for partial participation and network churn. 
The last global model $\bar\theta_i^{t-1}$ could, for example, date back to the penultimate aggregation step (i.e., to FL iteration $(t-2)$). 
To clarify that the last global model and last obtained smoothed delta might differ among peers, both are denoted using the peer indicator $i$. 
After initializing the noise-calibrating parameters $\sigma_b$ and $\sigma_\Delta$ using the number of participating aggregation peers $n_t$ and noise multiplier $\sigma_\mathrm{mult}$, peer $i$ prepares its DP-safe local model $\hat\theta_i^{t,0}$. 
This is done by computing the local model update vector $\Delta_i$, clipping, blurring, and smoothing it with factor $\beta$ to obtain $\bar\Delta_i^{t,0}$, and then finally deriving $\hat\theta_i^{t,0}$, where $\eta_u$ denotes the stepsize (we set $\beta = 0.9$ and $\eta_u = 0.1$). 
The noisy clipped local delta is denoted by $\widetilde{\Delta}_i$.
A binary indicator $b_i^{t,0}$ reveals whether peer $i$ has clipped its $\Delta_i$ to the clipping bound $C_t$. 
The squared noise calibration $\sigma_\Delta^2$ is rescaled by $n_t$ to account for noising local model deltas instead of their aggregated sum as~\citet{andrew21} do.

Over $G$ rounds of MAR, each group-based MAR aggregation step $\texttt{MAR}_g$ iteratively averages relevant peer information $\mathcal{P}_t$ from the set of participating aggregation peers $\mathcal{A}_t$ until each peer $i \in \mathcal{A}_t$ obtains: 
(i) a global state $(\theta^{t},\, m^{t})$,
(ii) a global clipping indicator $\bar{b}^t$, and 
(iii) a global smoothed delta $\bar{\Delta}^{t}$. 
The information peer $i$ has so far aggregated up to MAR round $g \in \{1,2,...,G\}$ of the current FL iteration $t$ is denoted as $(\hat\theta_i^{t,g},\, m_i^{t,g},\, b_i^{t,g},\, \bar{\Delta}_i^{t,g})$. 
A simple aggregation of binary indicators is not DP-safe as it reveals whether a peer $i$ has clipped its model update vector $\Delta_i$. 
To prevent this sensitive information leakage, a privacy-preserving mechanism (e.g., Secure Aggregation) has to be deployed for global binary indicator computation. 
When blurring the averaged binary indicator, sampled noise $\mathcal{N}(0,\,\sigma_b^2)$ is rescaled by the number of participating peers $n_t$, because we add noise to an average value and not to a sum as~\citet{andrew21} do. 
The global averaging of smoothed deltas from all participating peers $i \in \mathcal{A}_t$ ensures that during global model aggregation of the next FL iteration $(t+1)$, the privatized local delta $\widetilde{\Delta}_i$ is mixed with a privatized global momentum delta $\bar{\Delta}_i^t$ before being applied to the last global model $\bar\theta_i^{t}$. 
This yields variance reduction and global alignment when computing a DP-safe local model. 
Eventually, the clipping bound is updated to $C_{t+1}$, tracking a target quantile $\gamma$ of globally averaged clipping, where $\eta_C$ denotes the stepsize (we set $\gamma = 0.5$ and $\eta_C = 0.2$). 
After each aggregation, the DP-safe global model $\theta^t$ is stored as the peer’s 
last global model $\bar{\theta}_i^t$, to be used in its next global aggregation iteration; analogously for $\bar{\Delta}^t$.
\hw{We note that the local momentum vectors $m_i^{t}$ are not private as noise is applied only when each peer communicates their final model update for an aggregation round.}

% \fm{In~\Cref{algo:dpmarfedavg}, local momentum vectors $m_i^{t}$ are not privatized. 
% When aiming to preclude adversaries from, for example, inferring dominant gradient directions linked to a user’s data distribution from past local model updates, momentum vectors should also be clipped and blurred. 
% This would require a separate adaptive momentum vector clipping mechanism, which was not the scope of this work and is not necessarily required according to our evaluation objective. 
% Our experiments on DP aim to integrate DP into MAR-FL and to thereby investigate the impact of privacy preservation on model utility. 
% The major impact on model utility stems from privatizing model parameters: clipping and blurring the model parameters directly corrupts the weights used for prediction, the noise and bias are permanently embedded in the model itself, whereas perturbations to optimizer states only indirectly and transiently affect future updates.}

\section{Additional Experimental Details}

\subsection{Simulation Environment}\label{subsection:app_simenv}
We run all experiments on a high-performance computing (HPC) cluster using \texttt{Slurm} as the job scheduler. 
Each experiment runs on a single node with 4$\times$H100 GPUs, 768\,GB memory, and 96 CPU cores, reserving the entire node. 
After resource allocation, the job launches an \texttt{Enroot} runtime inside the allocation. 
The runtime is built from an \texttt{Enroot} SquashFS image created by importing the \texttt{NGC} container \texttt{nvcr.io/nvidia/pytorch:22.04-py3}. 
Inside the container we use \texttt{Python~3.8}, \texttt{PyTorch}, and our MAR-FL and baseline implementations.

% For each job, we allocate all 4 GPUs, the full memory, and 90 out of 96 CPU cores. 
% After resource allocation, each job launches an \texttt{Enroot} runtime inside the allocation.

\subsection{Implementation Details}\label{subsection:app_impdet}

\textbf{Process model.} 
We simulate peers as separate Python \texttt{multiprocessing} processes, each spawned by a dispatcher. 
Processes are created under a \texttt{spawn} context, assigned a unique peer ID, and pinned to specific CPU cores to simulate vCPUs. 
A shared \texttt{multiprocessing.Manager()} exposes two queues (task/results) and a shared dictionary for model exchange between the dispatcher and peers.

\textbf{Dispatcher.} 
A central dispatcher loop orchestrates FL iterations by:
(i) selecting participating peers for local updates and aggregation (modeling partial participation and churn),
(ii) enqueueing per-peer tasks (\texttt{update}, \texttt{aggregate}, \texttt{skip}, \texttt{shutdown}) on the task queue,
(iii) collecting results, logging timings, and monitoring communication volume,
(iv) performing early-stopping and robustness checks, and
(v) periodically clearing stale entries from the shared dictionary.

\textbf{Peer lifecycle.} 
Each peer process follows three steps:
(i) initialize a Hivemind DHT node to synchronize lightweight barriers and group-formation metadata (note that no model tensors are sent over the DHT),
(ii) load its local data partition (MNIST or 20NG) and the ML model (CNN or DistilBERT head), and
(iii) repeatedly execute tasks pulled from the task queue.

\textbf{Group formation and synchronization (MAR-FL).} 
At the beginning of the first MAR round, every peer initializes its group key.
In each MAR round, peers then:
(i) publish their presence via the DHT and collect peers with the same key,
(ii) enforce group symmetry by cross-checking gathered group members through DHT keys,
(iii) perform communication and aggregation within that group, and
(iv) update the group key via a deterministic schedule before the next round. 
To prevent repeatedly matching with the same peers, group key updates leverage each peer's chunk index. 
This procedure aligns with the MAR algorithm of~\citet{ryabinin21}.

\subsection{Experimental Setup}\label{subsection:app_expset}

\textbf{Datasets and models.}
To evaluate MAR-FL and all baselines on two distinct learning problems, we use one vision task (image classification) and one language task (text classification). 
For handwritten-digit recognition we employ a small two-block convolutional network with a compact multilayer-perceptron head that outputs class logits. 
MNIST images are loaded via \texttt{torchvision} and normalized in the usual way. 
For topic classification we use a lightweight classifier head on top of a frozen DistilBERT encoder~\citep{Sanh19}; the sequence representation is obtained from the classification token’s (CLS token) hidden state, and the head produces 20-way logits. 
Text is tokenized with a BERT-base uncased tokenizer and sequences are padded to a fixed length. 
The 20 Newsgroups dataset is loaded from Hugging Face Datasets (\texttt{SetFit/20\_newsgroups}). 

\textbf{FL baselines.}
We do not utilize Butterfly All-Reduce (BAR) as an additional P2P FL baseline. 
BAR aims to reduce total communication load by assigning disjoint parameter chunks to different peers and only partially aggregating at each node. 
Under heterogeneous participation or network churn this yields incomplete/partially aggregated models, where the network might be stalled waiting for entire chunks of the model architecture. 
BAR consequently requires peers to be totally reliable. 
Hence we compare MAR-FL against FedAvg, RDFL, and AR-FL, which better reflect the characteristics of aggregation relevant to FL.

\section{Additional Results}

\subsection{Qualitative Results between MAR-FL and our Baselines}\label{subsection:app_qrmarflbaselines}

\begin{figure}[t]%[13]{R}{\textwidth}
  \vspace{-0pt}
  \centering
  \begin{minipage}[t]{0.48\linewidth}
    \centering
    \captionsetup[subfigure]{justification=centering}
    \includegraphics[width=\linewidth]{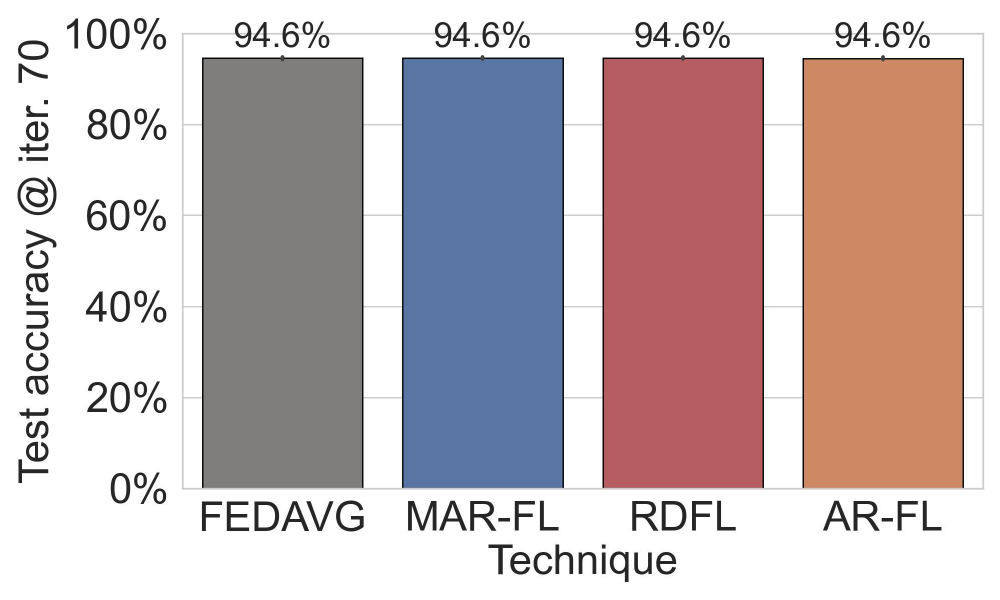}
    \subcaption{MNIST}
  \end{minipage}\hfill
  \begin{minipage}[t]{0.48\linewidth}
    \centering
    \captionsetup[subfigure]{justification=centering}
    \includegraphics[width=\linewidth]{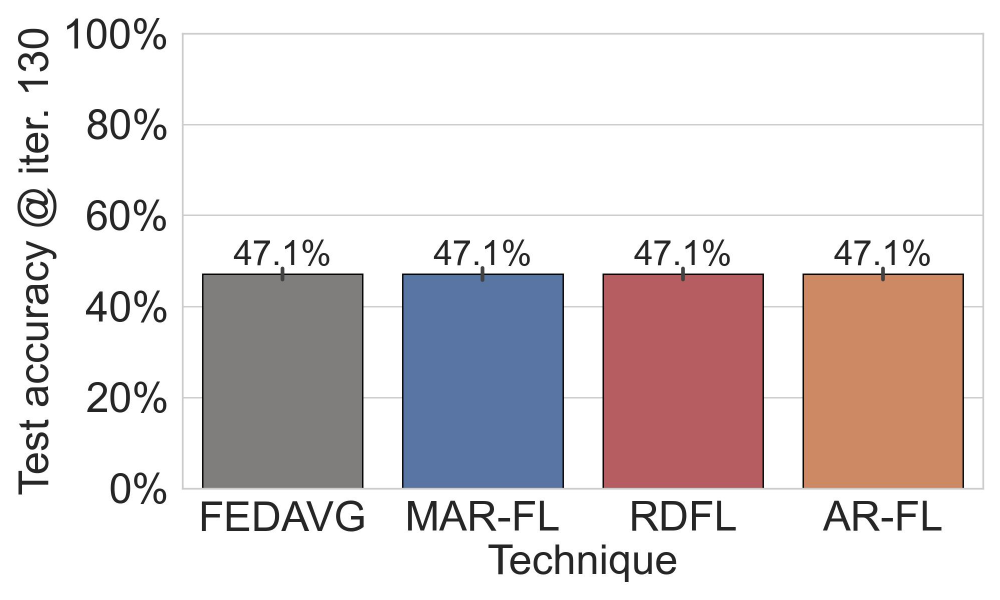}
    \subcaption{20NG}
  \end{minipage}\hfill
  \begin{minipage}[t]{0.48\linewidth}
    \centering
    \captionsetup[subfigure]{justification=centering}
    \includegraphics[width=\linewidth]{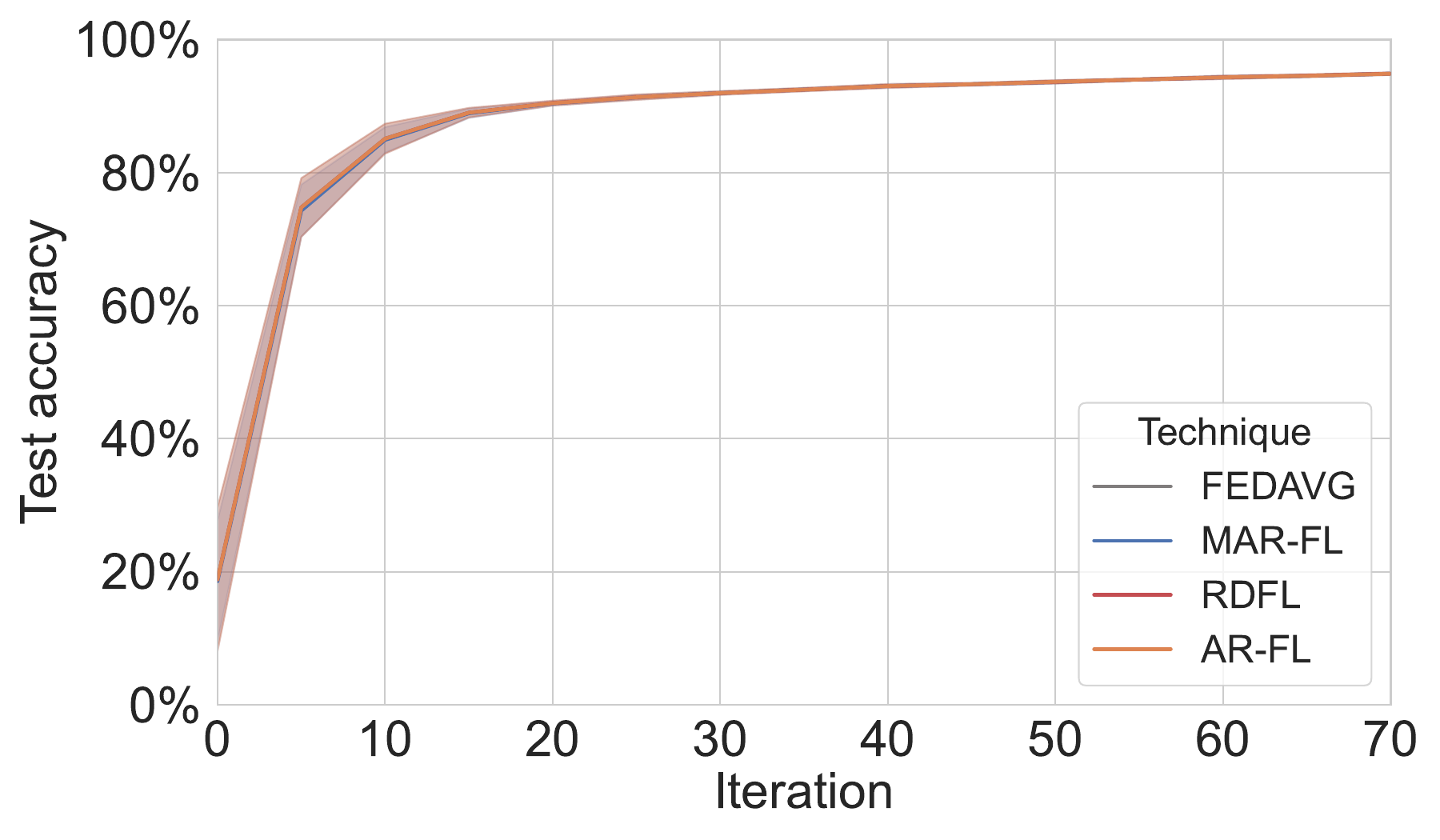}
    \subcaption{MNIST}
  \end{minipage}\hfill
  \begin{minipage}[t]{0.48\linewidth}
    \centering
    \captionsetup[subfigure]{justification=centering}
    \includegraphics[width=\linewidth]{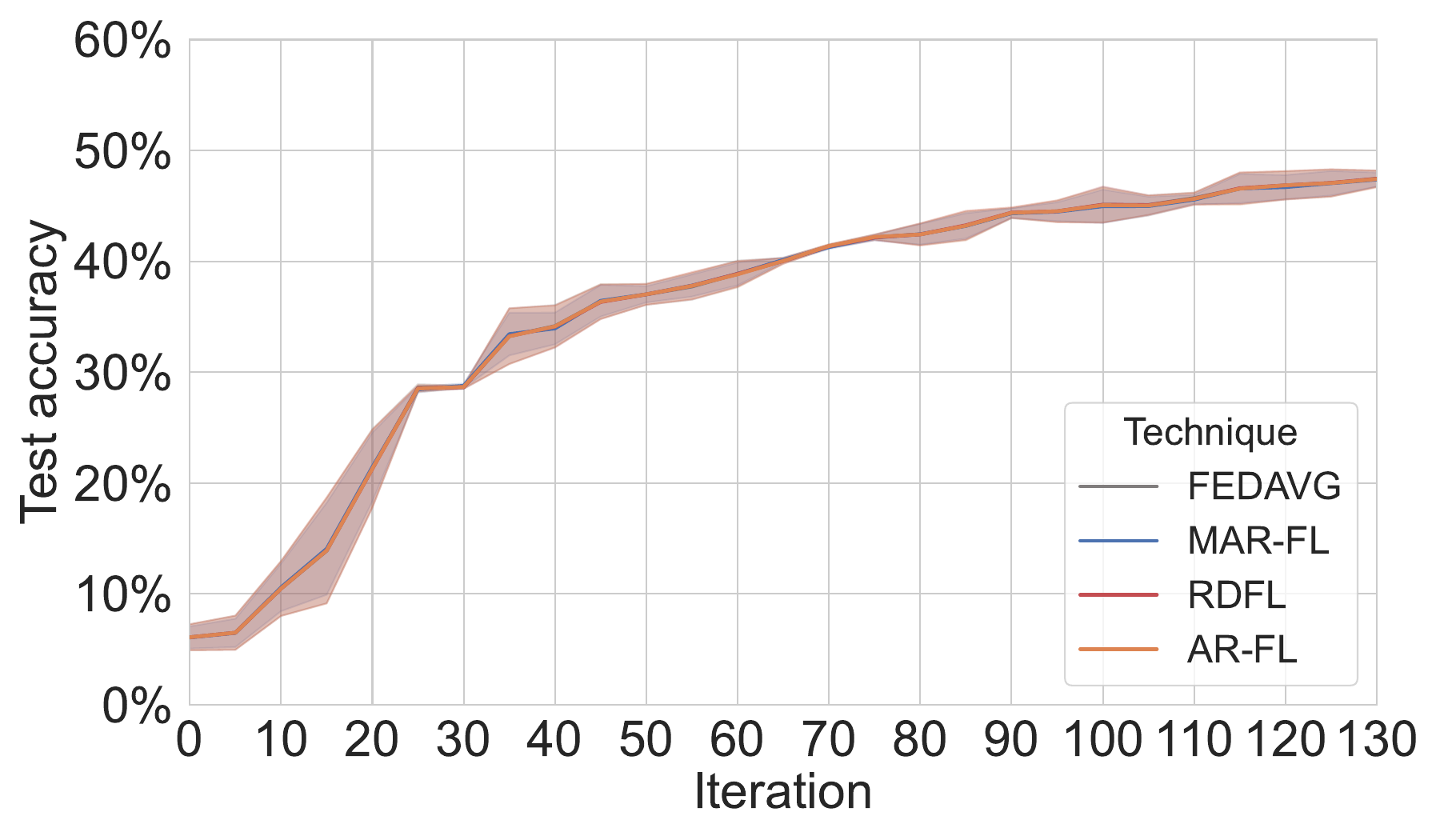}
    \subcaption{20NG}
  \end{minipage}
  \captionsetup{skip=5pt}
  \caption{MAR-FL yields the same test accuracy as client-server FedAvg and our P2P FL baselines.}
  \label{fig:e11_qualitative_identity}
  \vspace{-10pt}
\end{figure}

\textbf{Qualitative identity.}
On MNIST and 20NG, MAR-FL achieves the same training performance as client-server FedAvg and the two P2P FL baselines (see~\Cref{fig:e11_qualitative_identity}), as all four techniques yield identical global model averages under the given configurations.

\begin{figure}[t]%[13]{R}{\textwidth}
  \vspace{-0pt}
  \centering
  \begin{minipage}[t]{0.48\linewidth}
    \centering
    \captionsetup[subfigure]{justification=centering}
    \includegraphics[width=\linewidth]{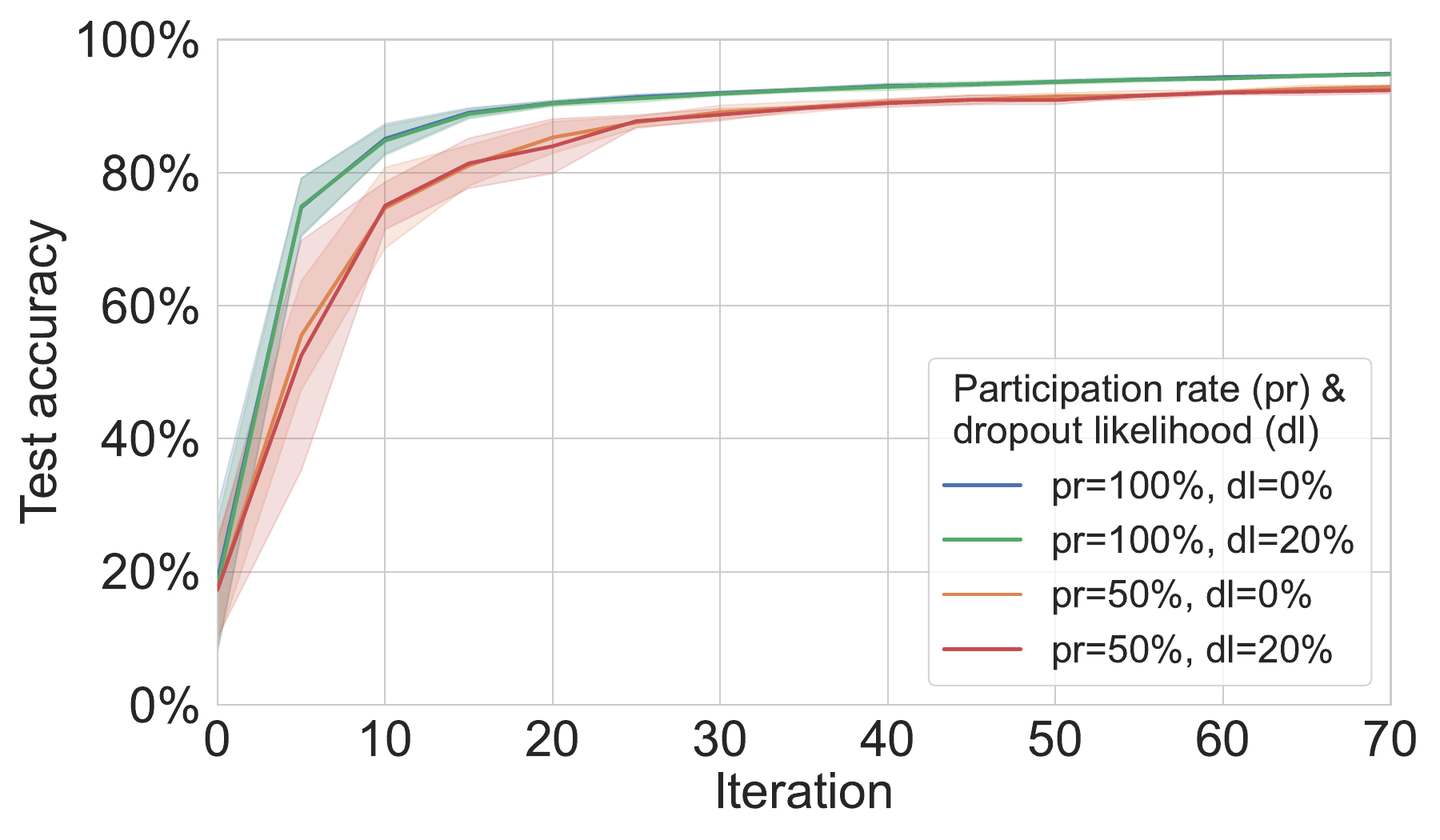}
    \subcaption{Model Performance for MAR-FL only}
  \end{minipage}\hfill
  \begin{minipage}[t]{0.43\linewidth}%48
    \centering
    \captionsetup[subfigure]{justification=centering}
    \includegraphics[width=\linewidth]{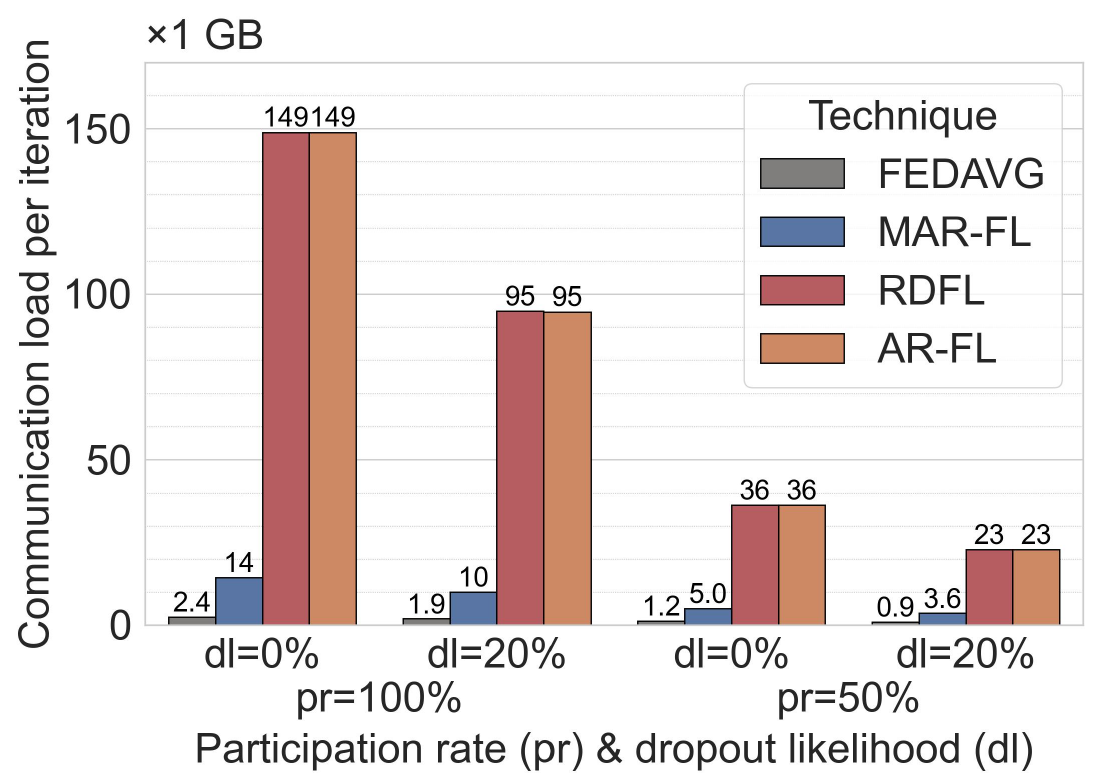}
    \subcaption{Communication Cost}
  \end{minipage}
  \captionsetup{skip=5pt}
  \caption{MAR-FL is affected by partial participation but resilient towards sudden dropouts. Plots show results on MNIST.}
  \label{fig:e4_partial_participation_mnist}
  \vspace{12pt}
  \begin{minipage}[t]{0.33\linewidth}
    \centering
    \captionsetup[subfigure]{justification=centering}
    \includegraphics[width=\linewidth]{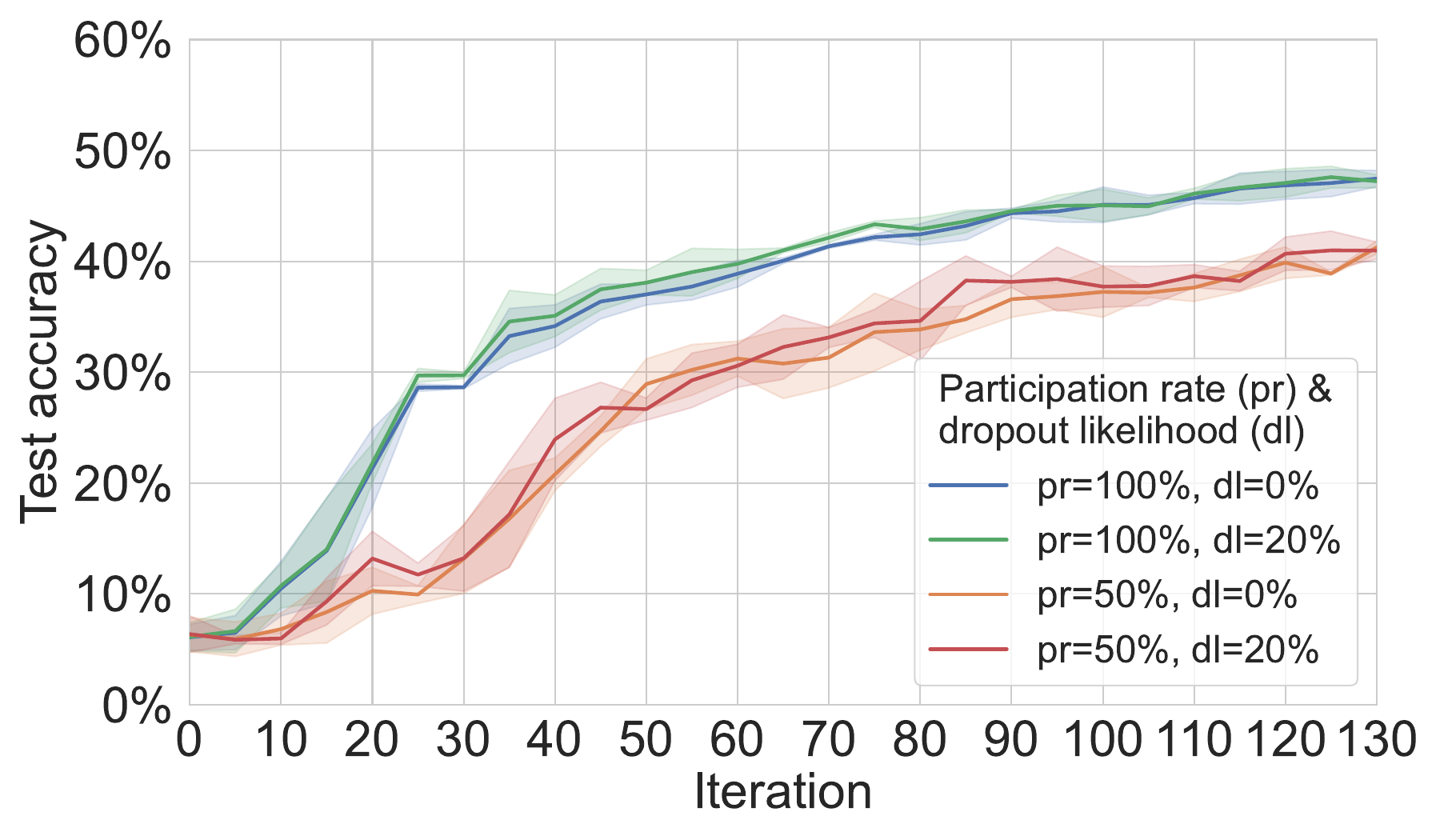}
    \subcaption{FEDAVG}
  \end{minipage}\hfill
  \begin{minipage}[t]{0.33\linewidth}
    \centering
    \captionsetup[subfigure]{justification=centering}
    \includegraphics[width=\linewidth]{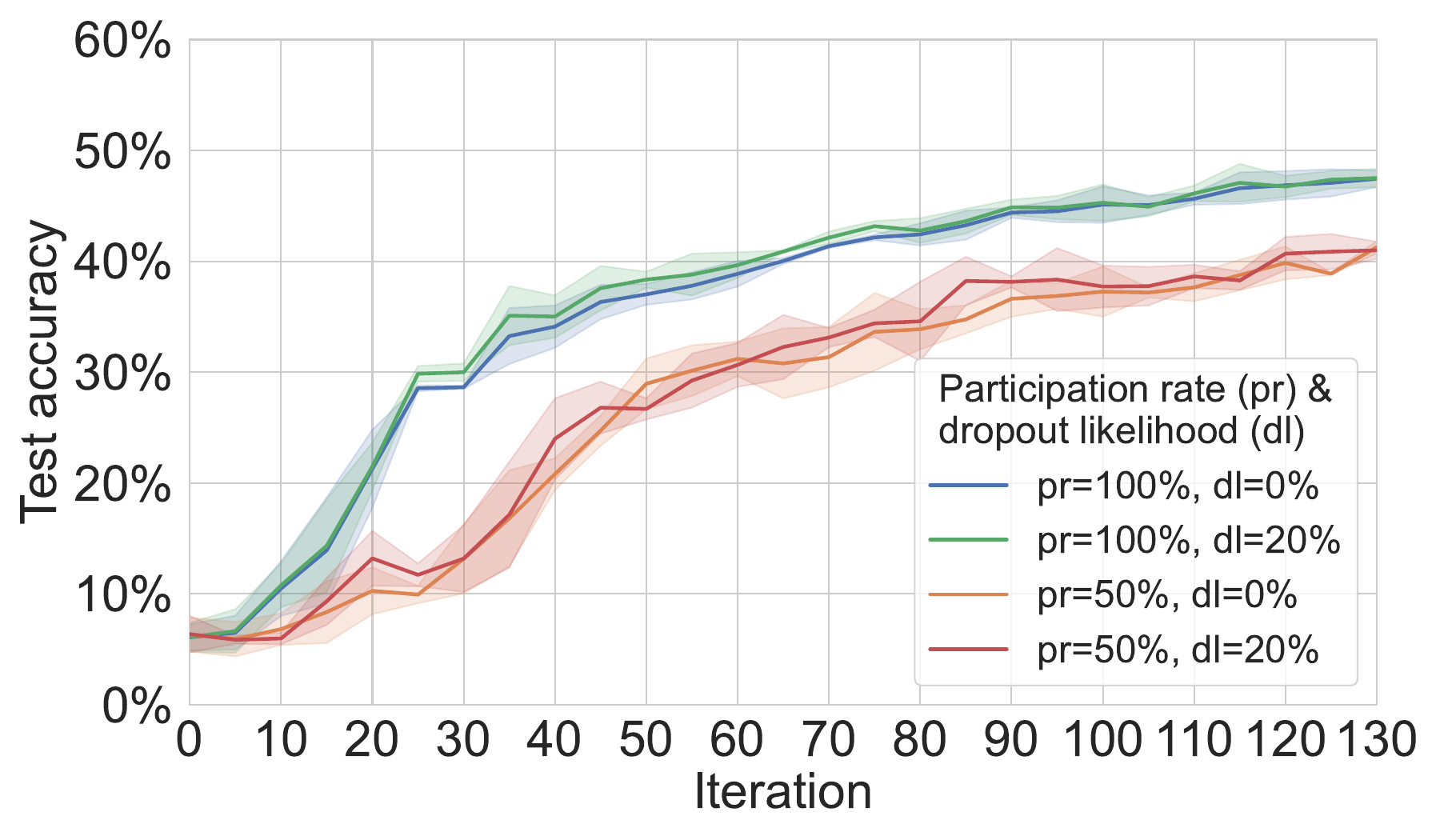}
    \subcaption{RDFL}
  \end{minipage}\hfill
  \begin{minipage}[t]{0.33\linewidth}
    \centering
    \captionsetup[subfigure]{justification=centering}
    \includegraphics[width=\linewidth]{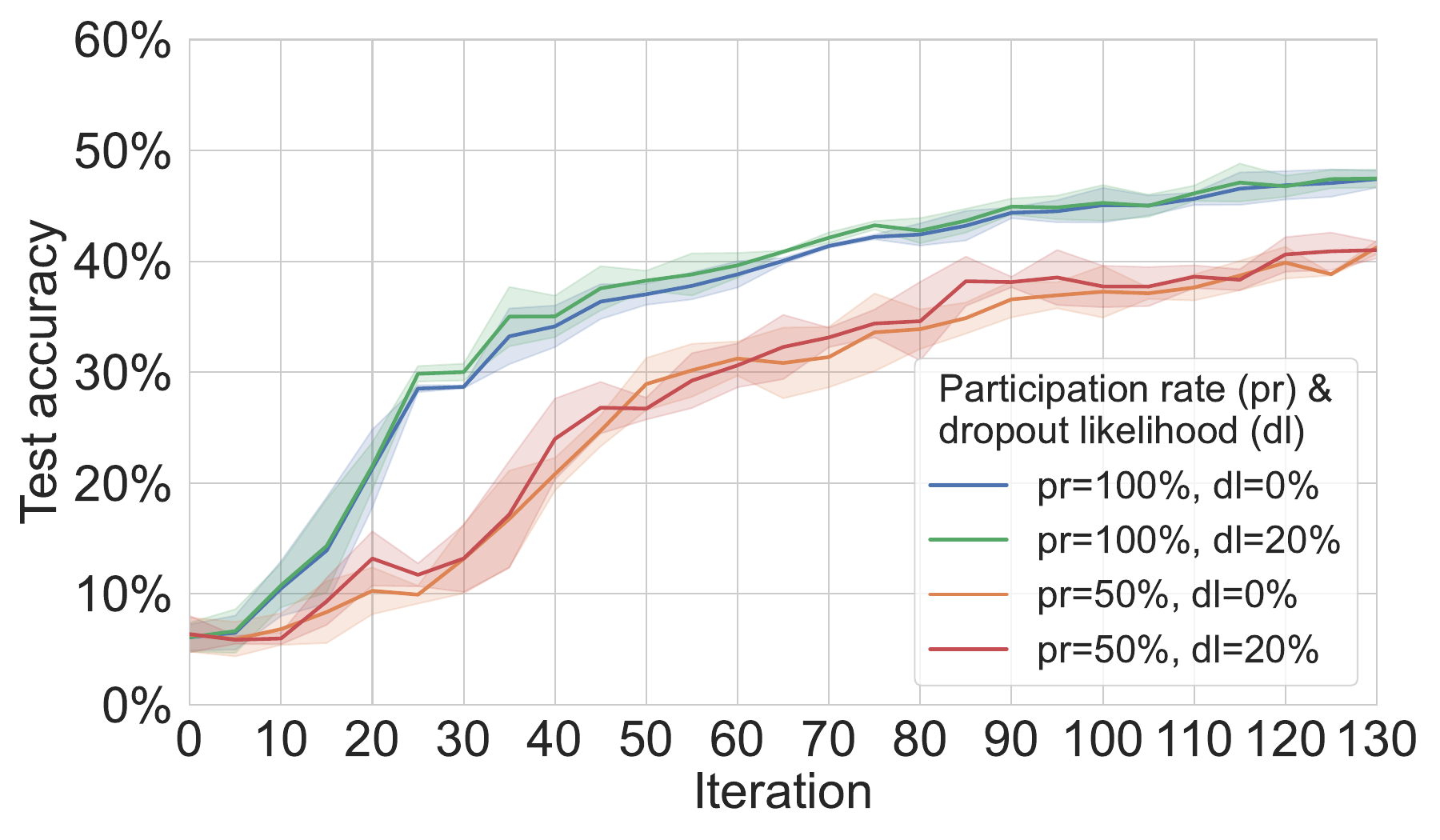}
    \subcaption{AR-FL}
  \end{minipage}
  \captionsetup{skip=5pt}
  \caption{Under partial participation and unreliable clients, FedAvg and our P2P FL baselines exhibit the same impact on training performance as shown for MAR-FL in~\Cref{fig:e4_partial_participation_news}. Plots show results on 20NG.}
  \label{fig:e43_partial_participation_baselines}
  \vspace{-10pt}
\end{figure}

\textbf{Partial participation.} 
On MNIST, MAR-FL incurs some loss in model utility under partial participation (see~\Cref{fig:e4_partial_participation_mnist}), though the degradation is milder than on 20NG. 
However, even with only $50\%$ peer participation and a $20\%$ dropout likelihood, 
MAR-FL remains more than $5\times$ as communication-efficient as our two P2P FL baselines. 
On 20NG, \Cref{fig:e43_partial_participation_baselines} shows that FedAvg and both P2P baselines degrade to a similar extent, consistent with the behavior observed for MAR-FL (see~\Cref{fig:e4_partial_participation_news}).

\subsection{Qualitative Results of MAR-FL}\label{subsection:app_qrmarfl}

\begin{figure}[t]%[12]{R}{0.80\textwidth}
  \vspace{-0pt}
  \centering
  \begin{minipage}[t]{0.48\linewidth}
    \centering
    \captionsetup[subfigure]{justification=centering}
    \includegraphics[width=\linewidth]{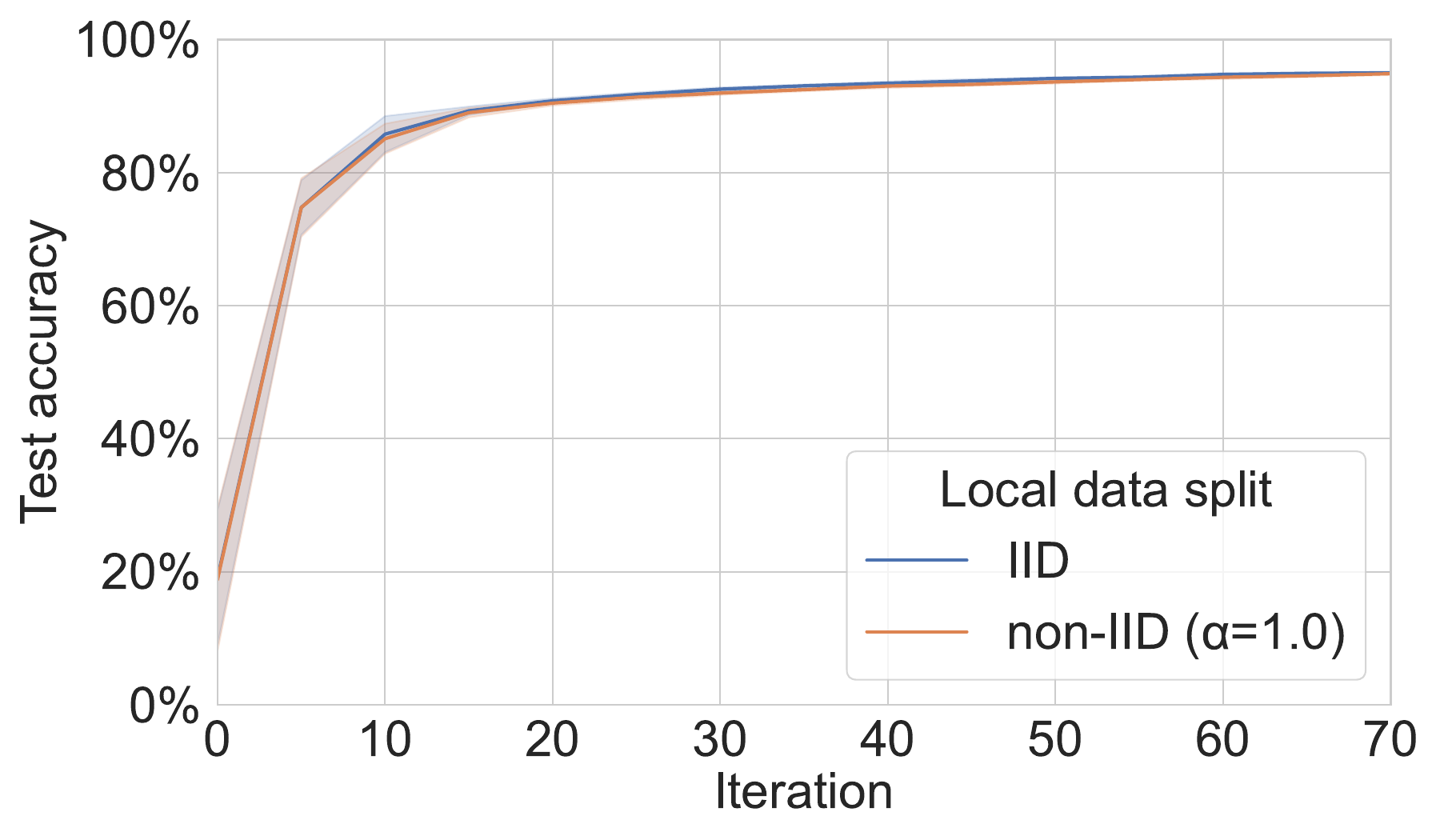}
    \subcaption{MNIST}
  \end{minipage}\hfill
  \begin{minipage}[t]{0.48\linewidth}
    \centering
    \captionsetup[subfigure]{justification=centering}
    \includegraphics[width=\linewidth]{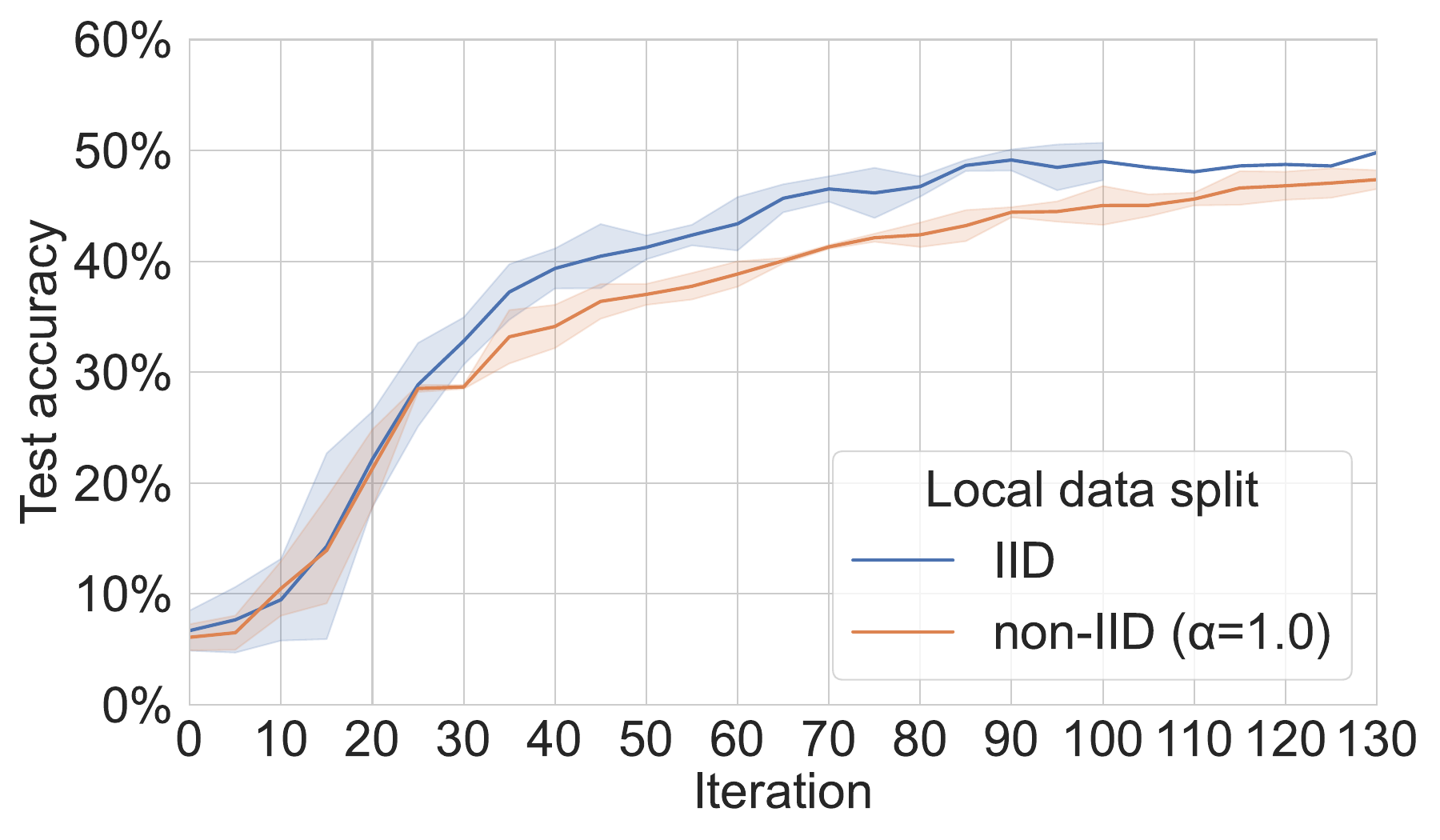}
    \subcaption{20NG}
  \end{minipage}
  \captionsetup{skip=5pt}
  \caption{Training performance of MAR-FL under i.i.d. and non-i.i.d. local data splits: performance on MNIST remains stable, whereas on 20NG non-i.i.d. splits lead to a noticeable degradation.}
  \label{fig:e13_iid_vs_niid}
  \vspace{-0pt}
\end{figure}

\textbf{Heterogeneous peer data.}
We employ Latent Dirichlet Allocation ($\alpha = 1.0$) to create non-i.i.d. local data splits among participating peers. 
While our simulation of real-world heterogeneous data distributions has no significant effect when training MAR-FL on MNIST, performance on 20NG is noticeably impaired compared to training with nearly i.i.d. local data splits (see~\Cref{fig:e13_iid_vs_niid}). 

\begin{figure}[t]%[12]{R}{0.38\textwidth}
  \vspace{-0pt}
  \centering
  \begin{minipage}[t]{0.48\linewidth}
      \centering
      \includegraphics[width=\linewidth]{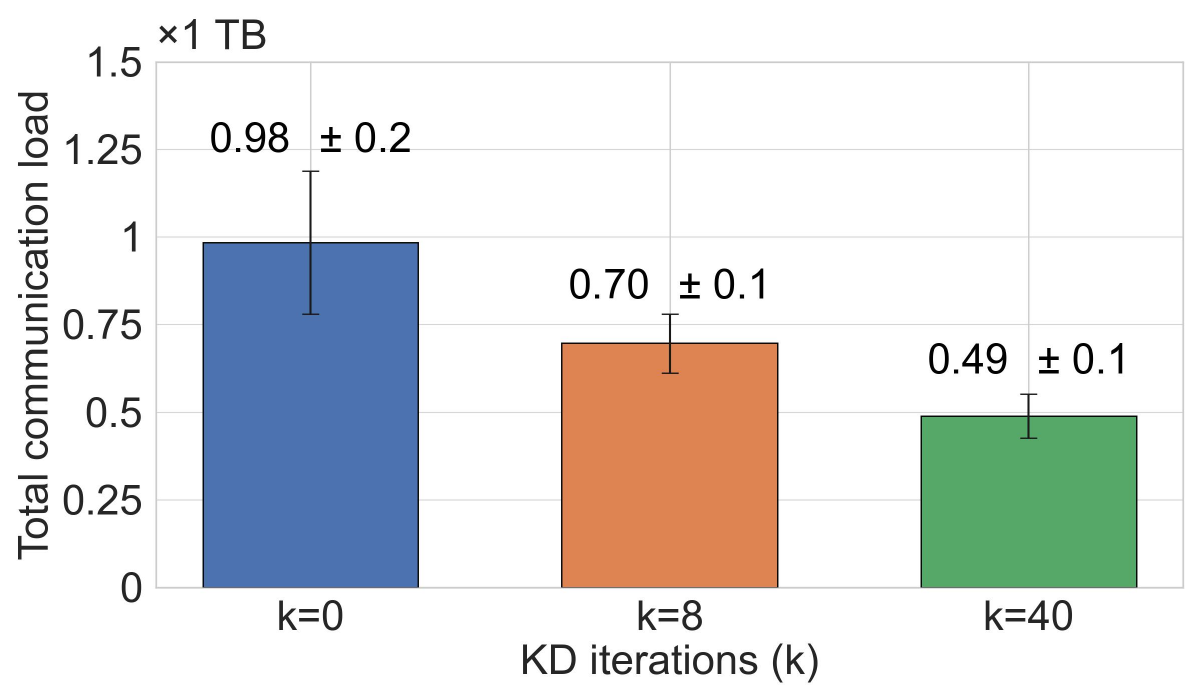}
      \captionsetup{skip=5pt}
      \caption{On MNIST, KD enables MAR-FL to reach a target accuracy of $95\%$ with substantially lower communication cost (up to $3\times$).}
      \label{fig:e22_kd_mnist}
  \end{minipage}\hfill
  \begin{minipage}[t]{0.48\linewidth}
      \centering
      \includegraphics[width=\linewidth]{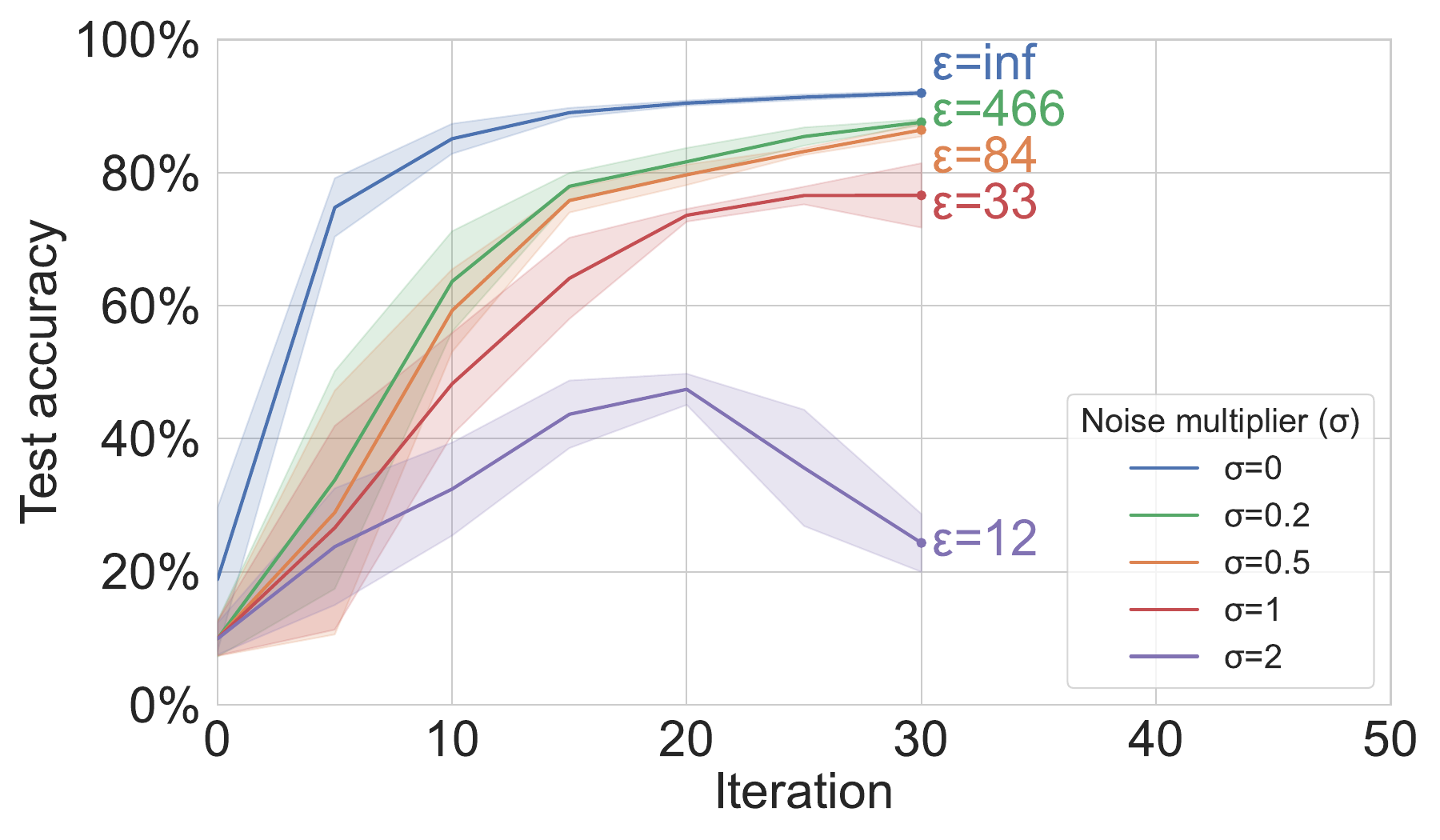}
      \captionsetup{skip=5pt}
      \caption{Integrating DP into MAR-FL results in the same performance degradation patterns as when applying DP to FedAvg. Plot shows results on MNIST.}
      \label{fig:e31_dp_mnist}
  \end{minipage}
  \vspace{-10pt}
\end{figure}

\textbf{Improving communication efficiency with MKD.}
As on 20NG, MKD also accelerates convergence for MAR-FL on MNIST, enabling a
target accuracy of $95\%$ to be reached with up to $3\times$ lower total communication
(see~\Cref{fig:e22_kd_mnist}), despite the increased per-iteration load from global aggregation. 
The number of KD iterations $k$ is chosen such that -- without data loader shuffling -- for $k=8$ on MNIST and $k=6$ on 20NG each peer processes its entire local dataset twice, while for $k=40$ on MNIST and $k=30$ on 20NG it is seen ten times. 

\textbf{Differentially private training.}
As observed for 20NG, increasing privatization for MAR-FL on MNIST (i.e., raising the noise multiplier value $\sigma$) eventually degrades training performance (see~\Cref{fig:e31_dp_mnist}). 

% The spent privacy budget of $\varepsilon = 33$ for DP with $\sigma = 1$ remains above common recommendations~\citep{wei20,andrew21}. 
% When MKD is combined with DP, it slightly accelerates convergence on MNIST but slows down training on 20NG (see~\Cref{fig:e32_dp_kd}). 
% The absence of the performance boost we obtained without activated DP can be attributed to several factors: students rely on privatized teachers whose soft targets are clipped and noisy, the DP-safe removal of the CE term weakens the learning effect, EMA smoothing further dilutes MKD’s signal, clipping bound updates adapt to binary indicators computed prior to applying MKD. 
% \textcolor{red}{KD was thereby parametrized conservatively with $k=8$ on MNIST and $k=6$ on 20NG.}

\begin{figure}[t]%[12]{R}{0.80\textwidth}
  \vspace{-245pt}
  \centering
  \begin{minipage}[t]{0.48\linewidth}
    \centering
    \captionsetup[subfigure]{justification=centering}
    \includegraphics[width=\linewidth]{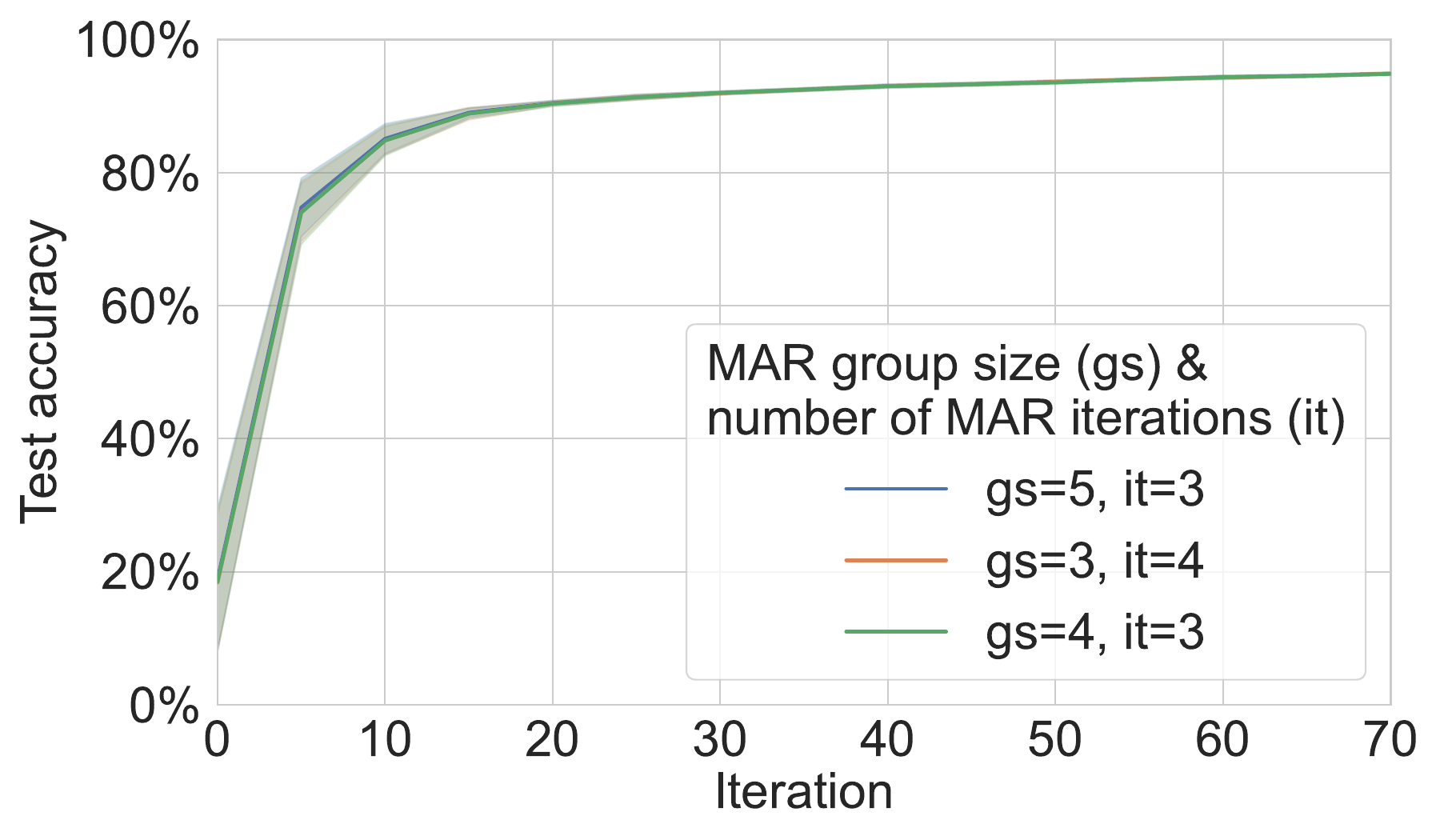}
    \subcaption{Model Performance on MNIST}
  \end{minipage}\hfill
  \begin{minipage}[t]{0.48\linewidth}
    \centering
    \captionsetup[subfigure]{justification=centering}
    \includegraphics[width=\linewidth]{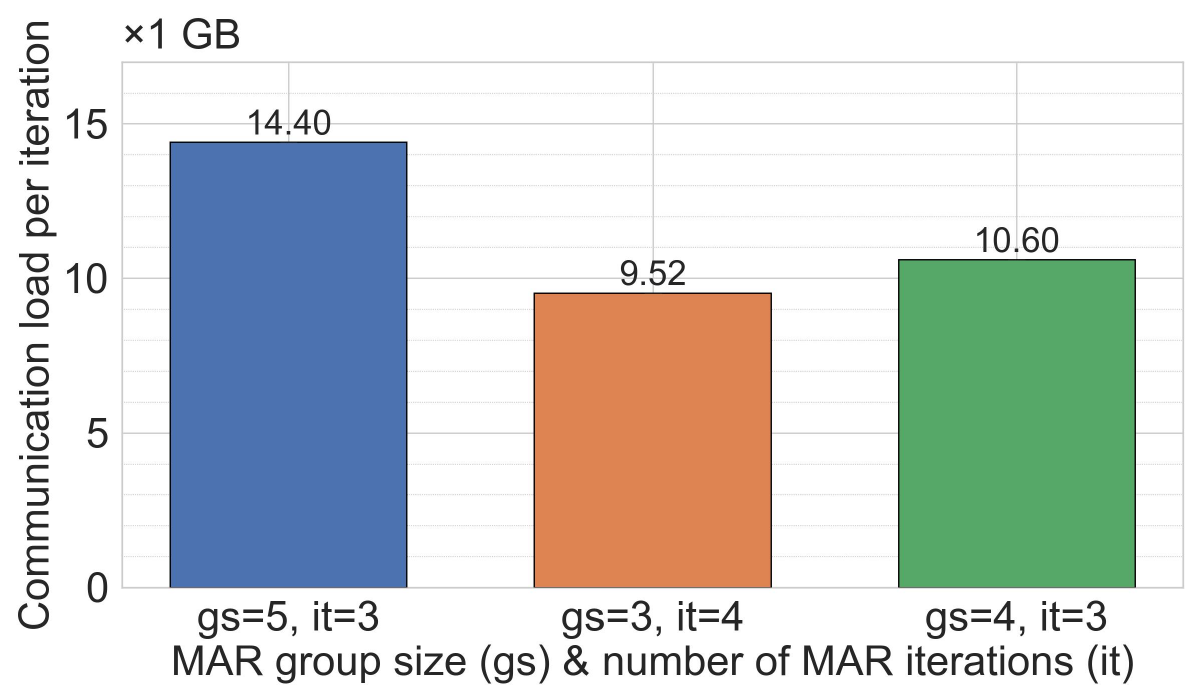}
    \subcaption{Communication Cost on MNIST}
  \end{minipage}
  \begin{minipage}[t]{0.48\linewidth}
    \centering
    \captionsetup[subfigure]{justification=centering}
    \includegraphics[width=\linewidth]{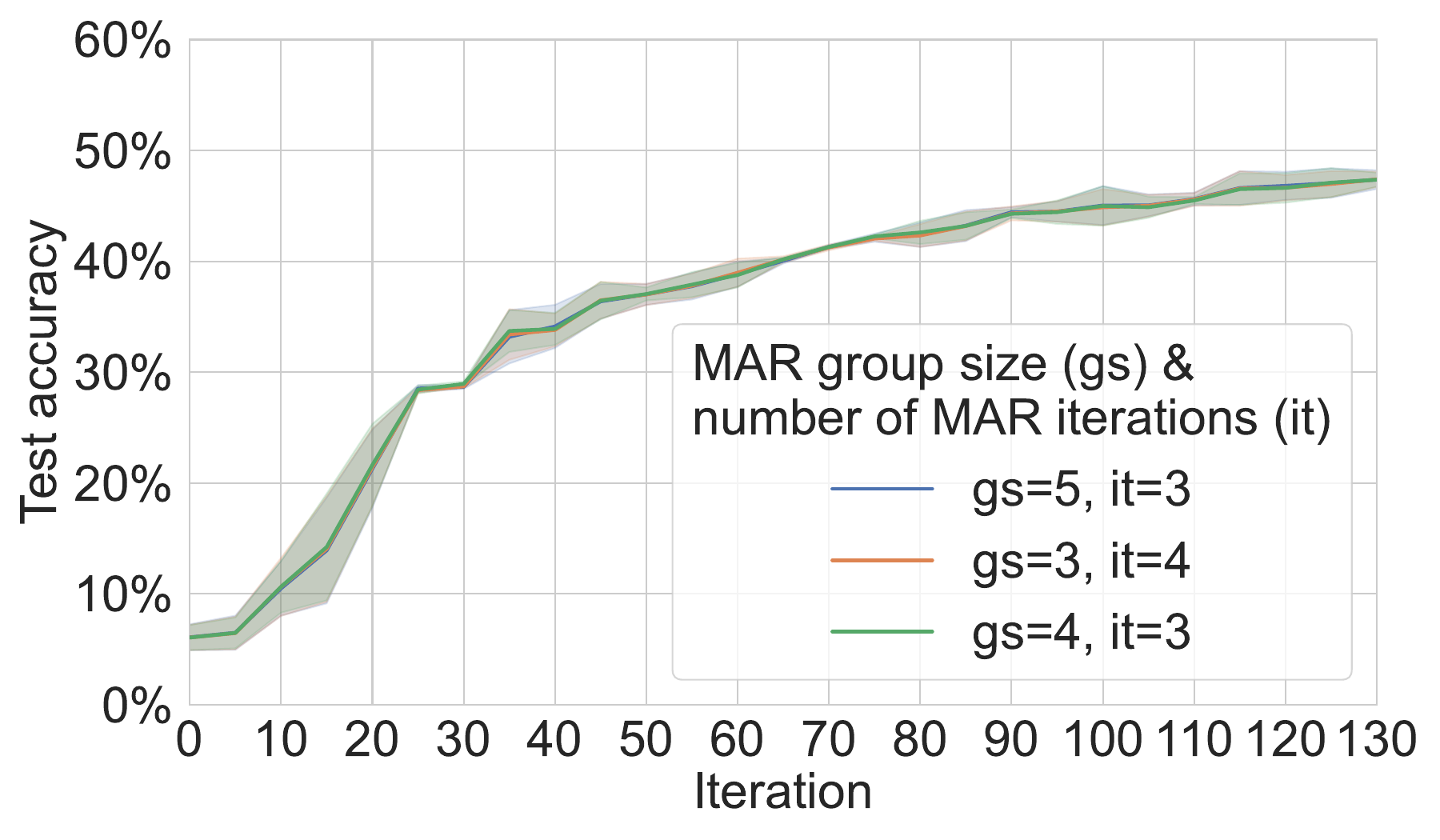}
    \subcaption{Model Performance on 20NG}
  \end{minipage}\hfill
  \begin{minipage}[t]{0.48\linewidth}
    \centering
    \captionsetup[subfigure]{justification=centering}
    \includegraphics[width=\linewidth]{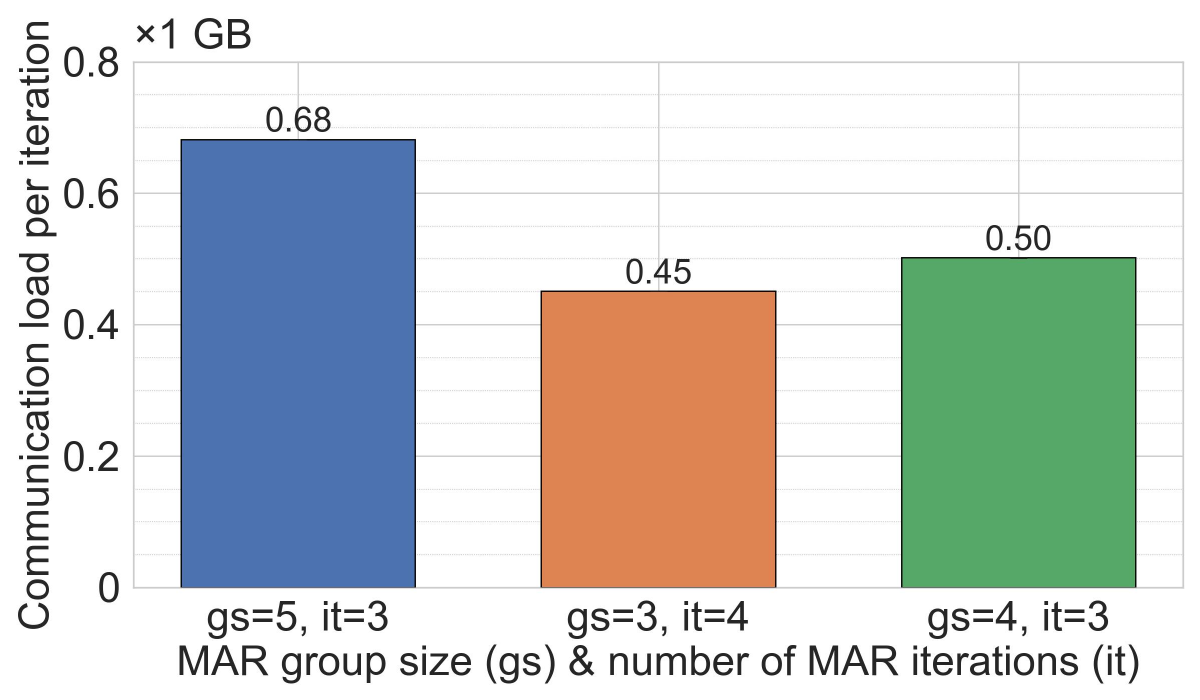}
    \subcaption{Communication Cost on 20NG}
  \end{minipage}
  \captionsetup{skip=5pt}
  \caption{On both ML tasks, appropriately configured approximate aggregation enables MAR-FL to preserve model utility while further reducing communication costs by up to $33\%$.}
  \label{fig:e5_approx_aggregation}
  \vspace{-10pt}
\end{figure}

\textbf{Leveraging approximate aggregation.}
\fm{As illustrated in~\Cref{fig:e5_approx_aggregation}, MAR-FL’s iterative group-based aggregation mechanism can be tuned to reduce communication while maintaining model utility. 
For example, with $125$ peers, MAR-FL achieves an exact global model average when using group size $5$ and $3$ MAR rounds, since $5^3 = 125$ (with group key dimension $d=3$). 
By relaxing these parameters (e.g., group size $3$ and $4$ MAR rounds), each iteration yields only approximate model averages. 
A well-designed group key update strategy is thus essential to closely approximate global averaging while minimizing the number of peer interactions per iteration. 
Over multiple iterations, these approximations converge to near-exact global averages, ensuring no substantial loss in model utility while significantly lowering communication cost.
In our experiments, communication was reduced by up to 33\% when using group size $3$ and $4$ MAR rounds with $125$ peers (group key dimension $d=4$).}

